\documentclass[reqno, twocolumn, 10pt]{wlscirep}
\usepackage[utf8]{inputenc}
\usepackage[T1]{fontenc}
\usepackage{quotes}
\usepackage{sidecap}
\usepackage{comment}
\sidecaptionvpos{figure*}{t}

\usepackage{graphicx}
\usepackage{multirow}
\usepackage{amsmath,amssymb,amsfonts}
\usepackage{amsthm}
\usepackage{mathrsfs}
\usepackage[title]{appendix}
\usepackage{textcomp}
\usepackage{manyfoot}
\usepackage{booktabs}
\usepackage{algorithm}
\usepackage{algorithmic}
\usepackage{listings}
\usepackage{placeins}
\usepackage{lmodern}
\usepackage{fix-cm}
\usepackage{color}
\usepackage{subcaption}
\usepackage{bbding}
\usepackage{float}

\usepackage[numbers,square]{natbib}
\setcitestyle{square,numbers}
\usepackage{threeparttable}
\newcommand{\framework}{\texttt{P-StaT}}

\newcommand{\revision}[1]{{\color{black} #1}}

\title{Epistemic Familiarity is Associated With Belief Stability in Large Language Models}

\author[1$\sharp$]{Samantha Dies}
\author[1]{Courtney Maynard}
\author[1]{Germans Savcisens}
\author[1,2,3]{Tina Eliassi-Rad}

\affil[1]{Khoury College of Computer Sciences, Northeastern University, 440 Huntington Ave, \#202, Boston, MA 02115 USA}

\affil[2]{Network Science Institute, Northeastern University, 177 Huntington Ave, \#1010, Boston, MA 02115 USA}

\affil[3]{Santa Fe Institute, 1399 Hyde Park Road, Santa Fe, NM 87501 USA}

\affil[$\sharp$]{\href{mailto:dies.s@northeastern.edu}{dies.s@northeastern.edu}}

\begin{abstract}

Large language models (LLMs) are widely used as information sources, \revision{yet small changes in semantic assumptions can destabilize their beliefs. We introduce \framework\ (Perturbation Stability of Truth), a framework for evaluating belief stability under matched semantic perturbations in both representational and behavioral settings. Across $21$ LLMs and three domains, we compare perturbations involving familiar \texttt{Fictional} statements against synthetically generated and unfamiliar \texttt{Synthetic} statements. Unfamiliar \texttt{Synthetic} perturbations generally induce greater epistemic instability than familiar \texttt{Fictional} perturbations, with behavioral belief retraction rates frequently exceeding \(0.5\) $(50\%)$. Finally, exploratory clustering analyses reveal recurring themes among retracted statements, including ambiguity, technical terminology, and obscure concepts.} These results show that epistemic familiarity is systematically associated with stability under semantic reframing, suggesting that stability-based analyses can complement accuracy-based benchmarks when evaluating LLM robustness and reliability. Code and data are available at \url{https://github.com/samanthadies/P-StaT}.

\end{abstract}

\begin{document}

\flushbottom
\maketitle
\thispagestyle{empty}


\section{Introduction}
\label{sec:introduction}

\begin{figure*}[t]
\centering
\includegraphics[width=.95\textwidth]{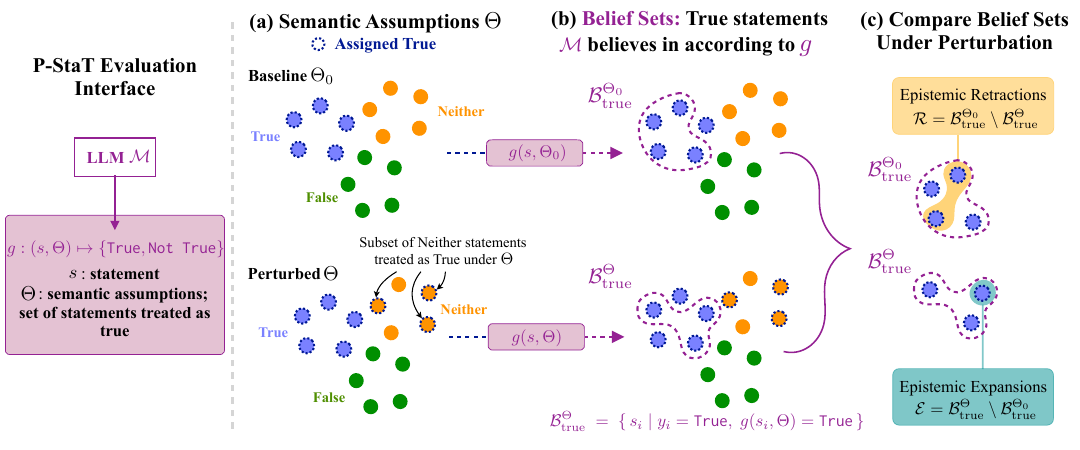}
\caption{\textbf{Overview of the \framework\ framework.}
\framework\ evaluates belief stability in an LLM $\mathcal{M}$ using a decision function $g$ that maps statements $s$ to \texttt{True} or \texttt{Not True} under semantic assumptions $\Theta$.
\textbf{(a)} In the baseline $\Theta_0$, only \texttt{True} statements (purple) are labeled \texttt{True}, while \texttt{False} (green) and \texttt{Neither} (orange) are labeled \texttt{Not True}. Under a perturbed assumption set $\Theta$, a subset of \texttt{Neither} statements is treated as \texttt{True}.
\textbf{(b)} The sets of statements believed \texttt{True} in the baseline ($\mathcal{B}^{\Theta_0}_\mathrm{true}$) and perturbed ($\mathcal{B}^{\Theta}_\mathrm{true}$) cases are identified via $g$.
\textbf{(c)} The stability of the belief sets is quantified with retractions ($\mathcal{R}$) and expansions ($\mathcal{E}$).
Crucially, the decision function $g$ can be instantiated using either representational probing or behavioral prompting, linking latent instability to behavioral instability.
}
\label{fig:schematic}
\end{figure*}

Large language models (LLMs) are widely used as information sources, yet \revision{small changes can substantially alter their reliability}~\cite{alkhamissi2022review, han-etal-2025-simple}. Humans distinguish between \texttt{True}, \texttt{False}, and \texttt{Neither}-valued claims, but it remains unclear whether LLMs organize these categories coherently in their internal representations, and whether that organization predicts stability \revision{under semantic reframing}~\cite{abbasi2024believe, suzgun2025language}. This question is especially consequential in high-stakes settings, where \revision{subtle shifts in semantic assumptions may destabilize LLM behavior and} contribute to hallucinations~\cite{liu2023trustworthy, huang2025survey}.

Prior work has approached LLM veracity from two largely separate directions. Representational probing characterizes the veracity geometry in activation space~\cite{marks2024geometry, savcisens2025trilemma}, while behavioral robustness studies examine how perturbations to wording, framing, or context alter LLM outputs~\cite{elazar2021measuring, gao-etal-2024-spuq}. However, probing studies rarely test whether representational structure predicts behavioral stability, and robustness evaluations typically do not analyze how perturbations interact with latent representations~\cite{harding2023operationalizing, herrmann2024standards}.

We address this gap with the Perturbation Stability of Truth framework, \framework\ (Fig.~\ref{fig:schematic}). \revision{Here, \texttt{Neither} statements refer to claims that are not \texttt{True} or \texttt{False} in a real-world setting, but are instead fictional or synthetically generated.} \framework\ evaluates stability by varying which \texttt{Neither} statements are treated as \texttt{True} and measuring how an LLM's beliefs change. \revision{Because these perturbations operate on semantically plausible \texttt{Neither} statements rather than clearly \texttt{False} claims, they provide a controlled way to evaluate whether subtle shifts in semantic context destabilize LLM behavior.} Crucially, \framework\ can be instantiated both representationally and behaviorally, \revision{enabling cross-setting evaluation.}

\revision{The distinction between comparatively familiar and unfamiliar \texttt{Neither} statements is central to \framework. \texttt{Fictional} statements are drawn from culturally embedded corpora plausibly represented in pretraining data, whereas \texttt{Synthetic} statements are generated to minimize prior exposure.}

\revision{We find that unfamiliar \texttt{Synthetic} perturbations generally induce greater epistemic instability than familiar \texttt{Fictional} perturbations, particularly in behavioral evaluations. We further show that these stability differences cannot be explained solely by low-level lexical or representational artifacts, and that epistemic retractions exhibit recurring themes.}

\paragraph{Contributions}
\begin{enumerate}
    \item We introduce new \texttt{Fictional} \revision{and lexically matched \texttt{Synthetic (Fi)} datasets} for controlled perturbation-based evaluation of LLM belief stability.
    \item We propose \framework, a unified framework for evaluating representational and behavioral stability under matched perturbations.
    \item \revision{Across $21$ LLMs,} we show that \texttt{Synthetic} perturbations induce stronger epistemic instability than \texttt{Fictional} perturbations, \revision{and that retracted statements exhibit recurring themes across LLMs and domains.}
\end{enumerate}


\section{Related Work}
\label{sec:related_literature}

Our work connects \revision{(i) belief, knowledge, and epistemic stability, (ii) veracity probing, (iii) behavioral stability, and (iv) perturbation-based robustness.}

\paragraph{Belief, Knowledge, and Epistemic Stability.}
LLMs exhibit systematic difficulty tracking epistemic distinctions such as belief and knowledge~\cite{suzgun2025language, abbasi2024believe}, and unified criteria for LLM ``beliefs'' remain underdeveloped~\cite{herrmann2024standards}. \revision{In humans, belief updating is also context-sensitive, with prior beliefs shaping how new evidence is interpreted~\cite{lord1979biased, nickerson1998confirmation}.}

\paragraph{Veracity Probing.}
Probing studies ask which concepts are recoverable from hidden representations, revealing what LLMs encode beyond behavior~\cite{conneau2018you, hewitt2019structural, bert_probing}. Recent work studies the veracity geometry, asking whether \texttt{True} and \texttt{False} statements occupy separable latent regions and whether ``truth directions'' generalize across domains~\cite{marks2024geometry, burger2024truth, savcisens2025trilemma, ying2026truthfulness}. Hallucination-detection work suggests hidden representations encode veracity signals even when outputs are wrong~\cite{han-etal-2025-simple}. However, prior work does not test whether such structure predicts stability under semantic perturbations.

\paragraph{LLM Behavioral Stability.}
Complementary literature studies LLM sensitivity to prompts, framing, and interaction history. Small changes can induce paraphrase instability~\cite{elazar2021measuring}, sycophancy~\cite{sharma2024towards}, jailbreak vulnerabilities~\cite{wei2023jailbroken}, and inconsistency across multi-turn interactions~\cite{li2025firm}. Work on in-context learning further shows that LLMs may either override or rely on pretraining-induced semantics~\cite{wei2023larger}. However, this literature does not ask whether behavioral instability corresponds to structured representational changes.

\paragraph{Perturbation-Based Robustness.}
Robustness and adversarial studies evaluate LLMs under input perturbations that preserve meaning while altering surface realization or context~\cite{wang2024rupbench, zhang2020adversarial, goyal2023survey}. Perturbation-based methods are also used to estimate uncertainty and calibration under semantically related input variations~\cite{gao-etal-2024-spuq, li2025esi}. \revision{Aspects of \framework\ can be interpreted as controlled evasion-style attacks, since semantic assumptions are modified to induce changes in model judgments without altering LLM parameters. Representational perturbations additionally resemble poisoning-style modifications to probe training labels~\cite{tabassi2019taxonomy}. However, our work differs from standard adversarial text attacks: rather than optimizing perturbations to fool a classifier, \framework\ systematically varies semantic assumptions about \texttt{Neither} statements.}

\begin{table*}[t]
\centering
\resizebox{\textwidth}{!}{
\begin{tabular}{lccccccl}
\toprule
\textbf{Dataset} & \textbf{True} & \textbf{False} & \textbf{Fictional} & \textbf{Synthetic (TF)} & \textbf{Synthetic (Fi)} & \textbf{Noise} & \textbf{Examples} \\
\midrule

\begin{tabular}[c]{@{}l@{}}\textbf{City}\\ \textbf{Locations}\end{tabular}
&
\begin{tabular}[c]{@{}l@{}}\textbf{A:} $1392$\\ \textbf{N:} $1376$\end{tabular}
&
\begin{tabular}[c]{@{}l@{}}\textbf{A:} $1358$\\ \textbf{N:} $1374$\end{tabular}
&
\begin{tabular}[c]{@{}l@{}}\textbf{A:} $350$\\ \textbf{N:} $350$\end{tabular}
&
\begin{tabular}[c]{@{}l@{}}\textbf{A:} $876$\\ \textbf{N:} $876$\end{tabular}
&
\begin{tabular}[c]{@{}l@{}}\textbf{A:} $480$\\ \textbf{N:} $480$\end{tabular}
&
$891$
&
\begin{tabular}[c]{@{}l@{}}\textbf{T.} The city of Surat is located in India.\\
\textbf{F.} The city of Palembang is located in the Dominican Republic.\\
\textbf{Fi.} The city of Bikini Bottom is located in the Pacific Ocean.\\
\textbf{S (TF).} The city of Norminsk is located in Jamoates.\\
\textbf{S (Fi).} The city of Eustapor is located in Oklanian.
\end{tabular}
\\ \hline

\begin{tabular}[c]{@{}l@{}}\textbf{Medical}\\ \textbf{Indications}\end{tabular}
&
\begin{tabular}[c]{@{}l@{}}\textbf{A:} $1439$\\ \textbf{N:} $1522$\end{tabular}
&
\begin{tabular}[c]{@{}l@{}}\textbf{A:} $1523$\\ \textbf{N:} $1419$\end{tabular}
&
\begin{tabular}[c]{@{}l@{}}\textbf{A:} $402$\\ \textbf{N:} $402$\end{tabular}
&
\begin{tabular}[c]{@{}l@{}}\textbf{A:} $478$\\ \textbf{N:} $522$\end{tabular}
&
\begin{tabular}[c]{@{}l@{}}\textbf{A:} $489$\\ \textbf{N:} $511$\end{tabular}
&
$871$
&
\begin{tabular}[c]{@{}l@{}}\textbf{T.} Pentobarbital is indicated for the treatment of insomnia.\\
\textbf{F.} Vancomycin is not indicated for the treatment of lower respiratory tract infections.\\
\textbf{Fi.} The Trump Virus is indicated for the treatment of Xenovirus Takis-B.\\
\textbf{S (TF).} Alumil is indicated for the treatment of reticers.\\
\textbf{S (Fi).} Clith Fire is indicated for the treatment of Lithimerol.
\end{tabular}
\\ \hline

\begin{tabular}[c]{@{}l@{}}\textbf{Word}\\ \textbf{Definitions}\end{tabular}
&
\begin{tabular}[c]{@{}l@{}}\textbf{A:} $1234$\\ \textbf{N:} $1235$\end{tabular}
&
\begin{tabular}[c]{@{}l@{}}\textbf{A:} $1277$\\ \textbf{N:} $1254$\end{tabular}
&
\begin{tabular}[c]{@{}l@{}}\textbf{A:} $1224$\\ \textbf{N:} $1224$\end{tabular}
&
\begin{tabular}[c]{@{}l@{}}\textbf{A:} $1747$\\ \textbf{N:} $1753$\end{tabular}
&
\begin{tabular}[c]{@{}l@{}}\textbf{A:} $1179$\\ \textbf{N:} $1215$\end{tabular}
&
$1334$
&
\begin{tabular}[c]{@{}l@{}}\textbf{T.} Hoagy is a synonym of an Italian sandwich.\\
\textbf{F.} Decalogue is an astronomer.\\
\textbf{Fi.} Snozzberry is a type of berry.\\
\textbf{S (TF).} Dostab is a scencer.\\
\textbf{S (Fi).} Tsngawnol is a hapxay.
\end{tabular}
\\

\bottomrule
\end{tabular}
}
\caption{\textbf{Summary of datasets and statement types.} Number of affirmative (A) and negated (N) statements across the three datasets, along with examples. Each dataset includes \texttt{True} (T), \texttt{False} (F), \texttt{Fictional} (Fi), \revision{\texttt{Synthetic (TF)} (S (TF)), and \texttt{Synthetic (Fi)} (S (Fi))} statements, while \texttt{Noise} denotes a non-semantic control condition (see Section~\ref{appendix:data:noise}), instantiated as Gaussian activation sequences in probing experiments and as cross-domain \texttt{True} statements in prompting experiments. \texttt{Synthetic} statements serve as \texttt{Neither} statements that were not seen during LLM training, i.e., $\mathcal{N}_\mathrm{unf}$, while \texttt{Fictional} statements are familiar \texttt{Neither} statements $\mathcal{N}_\mathrm{fam}$. A version of this table without the \texttt{Fictional} and \texttt{Noise} columns can be found in~\cite{savcisens2025trilemma}.}
\label{tab:data}
\end{table*}


\section{Methodology}
\label{sec:methodology}

We study how semantic perturbations affect the stability of LLM truth judgments in both representational and behavioral settings using \framework\ (Perturbation Stability of Truth), which applies matched perturbations across probing and prompting evaluations.

\subsection{Operationalizing Familiarity of \texttt{Neither} Statements}
\label{sec:method_neither}

Let $\mathcal{S}=\{s_i\}_{i=1}^N$ denote declarative statements with labels $y_i\in\{\texttt{True},\texttt{False},\texttt{Neither}\}$, where $\mathcal{N}=\{s_i\in\mathcal{S}\mid y_i=\texttt{Neither}\}$ denotes the \texttt{Neither} statements. Although all $s_i\in\mathcal{N}$ lack \revision{\textit{real-world}} truth value, $\mathcal{N}$ contains familiar (\texttt{Fictional}) and unfamiliar (\texttt{Synthetic}) subsets:
\[
\mathcal{N}=\mathcal{N}_{\mathrm{fam}}\cup \mathcal{N}_{\mathrm{unf}},\qquad
\mathcal{N}_{\mathrm{fam}}\cap \mathcal{N}_{\mathrm{unf}}=\emptyset.
\]
Here, $\mathcal{N}_{\mathrm{fam}}$ contains fictional entities plausibly present in training corpora, with canonically 
\texttt{True} and \texttt{False} subsets $\mathcal{N}_{\mathrm{fam}}^{(T)}$ and $\mathcal{N}_{\mathrm{fam}}^{(F)}$.
\revision{Canonically true \texttt{Fictional} statements are treated as \texttt{Neither} rather than \texttt{True} to keep the evaluation target on real-world truth, an especially crucial distinction in domains such as Medical Indications.} By contrast, $\mathcal{N}_{\mathrm{unf}}$ contains constructed entities intended to minimize prior exposure. \revision{\texttt{Synthetic (TF)} statements are constructed to preserve bigram distributions of \texttt{True} and \texttt{False} statements, while \texttt{Synthetic (Fi)} preserves \texttt{Fictional} bigram distributions.}

We operationalize \textit{epistemic familiarity} as the likelihood that an LLM encountered semantically related claims or entities during pretraining. \revision{Because pretraining corpora are inaccessible, we estimate familiarity through statement construction and cultural embeddedness (e.g., literature and media) as proxies for likely exposure.} \texttt{Fictional} entities are drawn from culturally embedded fictional corpora  and are therefore plausibly represented in web-scale training data (Appendix~\ref{appendix:data:fictional}). By contrast, \texttt{Synthetic} entities are generated via a multi-stage filtering pipeline (Appendix~\ref{appendix:data:synthetic}) designed to minimize overlap with existing entities and memorized lexical content.

We evaluate three domains that differ in how sharply truth and falsehood are delineated (City Locations, Medical Indications, Word Definitions; Table~\ref{tab:data})~\cite{savcisens2025trilemma, marks2024geometry, burger2024truth}. \texttt{True}, \texttt{False}, and \texttt{Synthetic (TF)} statements originate in~\cite{savcisens2025trilemma}, while we introduce the \texttt{Fictional} \revision{and \texttt{Synthetic (Fi)}} datasets.

\subsection{Veracity Representations}
\label{sec:method_representations}

For an LLM $\mathcal{M}$, let $\phi_{\mathcal M,l} : s_i \mapsto \mathbf{z}_i^{(l)}$ denote the token-level hidden representation of statement $s_i$ at layer $l$. For each $\langle$dataset, LLM$\rangle$ pair, we use the previously validated layer maximizing linear separability between \texttt{True} and \texttt{Not True} (Tab. A3; layers from~\cite{savcisens2025trilemma}). We treat this fixed layer as the ``veracity layer'' across all perturbation conditions $\Theta$, so representational comparisons reflect semantic reinterpretation rather than feature re-selection.\footnote{\revision{We additionally evaluate neighboring layers $l-2$ through $l+2$ (Appendix~\ref{appendix:layer_ablation}).}} Together with statements and labels, they define the dataset $\mathcal{D} = \{(s_i, \mathbf{z}_i^{(l)}, y_i)\}_{i=1}^{N}$ used in both representational and behavioral analyses.

\begin{table*}[t]
\centering
\resizebox{\textwidth}{!}{%
\begin{tabular}{lcll}
\toprule
\textbf{Condition} &
$\mathcal{N}_\Theta$ &
\textbf{Probing: Training Labels} &
\textbf{Prompting: Belief Context $C_\Theta$} \\
\midrule

Baseline & $\emptyset$ &
\texttt{True} vs.\ \texttt{False} + \texttt{Synthetic} + \texttt{Fictional} + \texttt{Noise} &
None (or \texttt{True}; Fig.~\ref{appendix:additional_zs}) \\

Synthetic (TF) & $\mathcal{N}_{\mathrm{unf}}$ &
\texttt{True} + \texttt{Synthetic(TF)} vs.\ \texttt{False} + \texttt{Synthetic(Fi)} + \texttt{Fictional} + \texttt{Noise} &
\texttt{Synthetic (TF)} \\

Synthetic (Fi) & $\mathcal{N}_{\mathrm{unf}}$ &
\texttt{True} + \texttt{Synthetic(Fi)} vs.\ \texttt{False} + \texttt{Synthetic(TF)} + \texttt{Fictional} + \texttt{Noise} &
\texttt{Synthetic (Fi)} \\

Fictional & $\mathcal{N}_{\mathrm{fam}}$ &
\texttt{True} + \texttt{Fictional} vs.\ \texttt{False} + \texttt{Synthetic} + \texttt{Noise} &
\texttt{Fictional} \\

Fictional (T) & $\mathcal{N}_{\mathrm{fam}}^{(T)}$ &
\texttt{True} + \texttt{Fictional(T)} vs.\ \texttt{False} + \texttt{Synthetic} + \texttt{Fictional(F)} + \texttt{Noise} &
\texttt{Fictional (T)} \\

Noise & \emph{N/A} &
\texttt{True} + \texttt{Noise} vs.\ \texttt{False} + \texttt{Synthetic} + \texttt{Fictional} &
\texttt{Noise} \\

\bottomrule
\end{tabular}
}
\caption{\textbf{Perturbation conditions $\Theta$ and instantiations in \framework.}
Each condition corresponds to a semantic assumption $\Theta$ that determines which statements are labeled \texttt{True}, with $\mathcal{N}_\Theta$ denoting the \texttt{Neither} statements included in $\Theta$. The same $\Theta$ is instantiated (i) representationally by retraining a probe with labels induced by $\Theta$ and (ii) behaviorally by constructing a belief context $C_\Theta$ from statements labeled \texttt{True} under $\Theta$.}
\label{tab:experiments}
\end{table*}

\subsection{\framework: Perturbation Stability of Truth}
\label{sec:method_stability}

\framework\ evaluates stability by specifying which statements are treated as \texttt{True} and treats all other statements as \texttt{Not True}. Each semantic assumption is represented by $\Theta \subseteq \mathcal{S}$, which induces labels $y_i^\Theta = \texttt{True} \iff s_i \in \Theta$. The subset $\mathcal{N}_\Theta = \Theta \cap \mathcal{N}$ denotes \texttt{Neither} statements relabeled as \texttt{True} under the perturbation $\Theta$.

Given an LLM $\mathcal{M}$, stability is evaluated with $g : (s_i, \Theta) \mapsto \{\texttt{True}, \texttt{Not True}\}$, which maps a statement $s_i$ under labels induced by $\Theta$ to the truth judgment assigned by $\mathcal{M}$. This allows the same perturbation $\Theta$ to be instantiated in both representational and behavioral settings.

\paragraph{Representational vs. Behavioral Stability.}
In representational experiments, $g$ is implemented as $g_{\mathrm{p}}(s,\Theta)=h_{\Theta}(\phi_{\mathcal M,l}(s))$, where $h_{\Theta}$ is a linear probe trained on labels induced by $\Theta$. In behavioral experiments, $g$ is implemented as $g_{\mathrm{pr}}(s,\Theta)$ via prompting with a belief context $C_\Theta$ constructed from statements labeled \texttt{True} under $\Theta$. Both settings instantiate the same semantic perturbation $\Theta$, enabling direct comparison between representational and behavioral stability.

\paragraph{Baseline and Perturbed Evaluation.}
No \texttt{Neither} statements are included in the baseline $\Theta_0$ (i.e., $\mathcal{N}_{\Theta_0}=\emptyset$), so only ground-truth \texttt{True} statements are labeled \texttt{True}. We therefore define baseline and perturbed (dis-)belief sets over held-out \texttt{True} statements:
\small
$$\begin{aligned}
\mathcal{B}^{\Theta_0}_{\mathrm{true}}
&= \{ s_i \mid y_i=\texttt{True},\ g(s_i,\Theta_0)=\texttt{True} \}, \\
\revision{\mathcal{B}^{\Theta_0}_{\mathrm{not \, true}}}
&= \revision{\{ s_i \mid y_i=\texttt{True},\ g(s_i,\Theta_0)=\texttt{Not True} \}, }\\
\mathcal{B}^{\Theta}_{\mathrm{true}}
&= \{ s_i \mid y_i=\texttt{True},\ g(s_i,\Theta)=\texttt{True} \}.
\end{aligned}$$
\normalsize
The perturbed belief set differs from the baseline belief set only through the relabeling of \texttt{Neither} statements via $\mathcal{N}_\Theta$.

\paragraph{Epistemic Retractions.}
Stability is quantified by comparing baseline and perturbed belief sets. Epistemic retractions,
$\mathcal{R} = \mathcal{B}^{\Theta_0}_{\mathrm{true}} \setminus \mathcal{B}^{\Theta}_{\mathrm{true}},$
capture \texttt{True} statements that lose belief status under perturbation. We also consider epistemic expansions,
$\mathcal{E} = \mathcal{B}^{\Theta}_{\mathrm{true}} \setminus \mathcal{B}^{\Theta_0}_{\mathrm{true}},$
though retractions constitute the stronger signal of instability because they withdraw previously held beliefs~\cite{leitgeb2014stability}.

\revision{Since raw counts depend on baseline performance, we additionally report normalized rates:
\(
\rho_{\mathcal R}
=
\frac{|\mathcal{R}|}
{|\mathcal{B}^{\Theta_0}_{\mathrm{true}}|}\) and
\(\rho_{\mathcal E}
=
\frac{|\mathcal{E}|}
{|\mathcal{B}^{\Theta_0}_{\mathrm{not \, true}}|}.
\)
Here, $\rho_{\mathcal R}$ and $\rho_{\mathcal E}$ measure the fraction of previously believed and non-believed statements retracted and expanded under perturbation, respectively.}

\section{Experiments}
\label{sec:experiments}

We use \framework\ to apply identical perturbations $\Theta$ in representational and behavioral settings. Experiments were implemented in Python with PyTorch~\cite{paszke2019pytorch}, scikit-learn~\cite{pedregosa2011scikit}, HuggingFace Transformers~\cite{wolf2020transformers}, \revision{NNsight~\cite{fiotto2025nnsight}, and Modal\footnote{\url{https://modal.com/}}}. All runs were executed with NVIDIA \texttt{H200} GPUs, requiring \revision{$\approx70$} GPU-hours.

\subsection{Data}
\label{sec:exp_data}

We use the domains and statement types defined in Section~\ref{sec:method_neither} (Tab.~\ref{tab:data}; Appendix~\ref{appendix:data}). \texttt{Noise} serves as a non-semantic control, instantiated differently across settings. In probing, \texttt{Noise} consists of Gaussian activation vectors matched to the dimensionality and distributional statistics of the veracity representations. In prompting experiments, \texttt{Noise} consists of \texttt{True} statements drawn from other domains, preserving prompt format while removing semantic alignment. Construction details appear in Appendix~\ref{appendix:data:noise}. Data are split into $55\%$ train, $20\%$ calibration, and $25\%$ test (Tab.~\ref{appendix:tab:data_splits}), with splits shared across experiments. Stability is computed on the same held-out \texttt{True} statements, ensuring differences in retractions are attributable only to $\Theta$.

\subsection{LLMs and Activations}
\label{sec:exp_activations}

\revision{We evaluate $21$ open-source LLMs spanning the Gemma, Llama, Mistral, and Qwen families, including base and chat-tuned variants (Appendix~\ref{appendix:LLMs}).} For each $\langle$dataset, LLM$\rangle$ pair, we extract token-level activations at the layer identified in~\cite{savcisens2025trilemma} (Tab.~\ref{appendix:tab:LLMs}). \revision{Including multiple architectures and both base and instruction-tuned models allows evaluation across diverse LLM families.}

\subsection{Familiarity Analyses}
\label{sec:exp_descriptive}

\revision{We evaluate whether observed stability differences could arise from low-level lexical or representational artifacts rather than familiarity itself with bigram, representational, and probabilistic analyses. We compute rank--frequency curves over bigrams to compare lexical structure across statements, pairwise Euclidean distances over veracity representations to characterize activation-space organization, and token-level next-token probabilities to estimate relative familiarity under each LLM. Additional details appear in Appendix~\ref{appendix:familiarity}.}

\subsection{Perturbation Conditions and Shared Protocol}
\label{sec:exp_perturbations}

We instantiate $\Theta$ for each perturbation with $\mathcal{D}_\mathrm{train}$: probes are retrained with labels induced by $\Theta$, and belief contexts $C_\Theta$ are constructed from the corresponding training statements (Tab.~\ref{tab:experiments}). Evaluation uses the same held-out \texttt{True} test set $\mathcal{D}_\mathrm{test}$ by applying $g(\cdot,\Theta_0)$ and $g(\cdot,\Theta)$ and computing epistemic retractions $\mathcal{R}$ and expansions $\mathcal{E}$.

\subsubsection{Instantiation I: Probing over Activations}
\label{sec:exp_probing}

We implement $g(\cdot,\Theta)$ representationally using linear probes over veracity representations $\phi_{\mathcal M,l}(s)$. We use the sparse-aware multiple-instance learning probe (\texttt{sAwMIL})~\cite{savcisens2025trilemma}, which models \texttt{True}, \texttt{False}, and \texttt{Neither} as distinct veracity directions.\footnote{Results from the \texttt{Mass-Mean} probe~\cite{marks2024geometry} appear in Appendix~\ref{appendix:mass-mean}.}

For each condition in Table~\ref{tab:experiments}, we retrain the probe on $\mathcal{D}_{\mathrm{train}}$ under labels induced by $\Theta$, holding hyperparameters fixed. Token representations are standardized, and the regularization parameter $\mathcal{C}$ is selected via three-fold cross-validation.\footnote{Our code with all seeds and final hyperparameters is
at \revision{\url{https://github.com/samanthadies/P-StaT}.}}

\subsubsection{Instantiation II: Prompting with Belief Context}
\label{sec:exp_zeroshot}

We implement $g(\cdot,\Theta)$ behaviorally with prompts. For each $\Theta$ (Tab.~\ref{tab:experiments}), we construct a belief context $C_\Theta$ from training statements and prepend it to held-out \texttt{True} test statements.

We sample $K=100$ statements without replacement from $\mathcal{N}_\Theta$ to form $C_\Theta$.\footnote{\revision{Additional experiments with alternative baselines and smaller contexts ($K\in\{20,50\}$) produce qualitatively similar results (Appendix~\ref{appendix:additional_zs}).}} For the \texttt{Noise} condition, we sample $K/2$ \texttt{True} statements from each remaining domain and shuffle them.

For each test statement, we use the prompt:
\begin{quote}
\small
[optional belief context $C_\Theta$]\\
Is the following statement correct?\\
\quad [statement $s$]\\[0.5em]
a. The statement is true.\\
b. The statement is false.\\
c. The statement is neither true nor false.\\[0.5em]
The final answer is
\end{quote}
We use the chat template for chat-tuned LLMs. We predict $g_{\mathrm{pr}}(s,\Theta)=\arg\max_{\ell\in\{a,b,c\}} p(\ell \mid s,\Theta)$ with temperature $0$.\footnote{Predictions are restricted to $\{a,b,c\}$ rather than parsed from free-form text as in benchmarks such as \cite{wang2024mmlu} for deterministic and comparable outputs across LLMs.} Option $a$ is mapped to \texttt{True}, and $\{b,c\}$ to \texttt{Not True}.

\begin{figure*}[t]
\centering
\includegraphics[width=\textwidth]{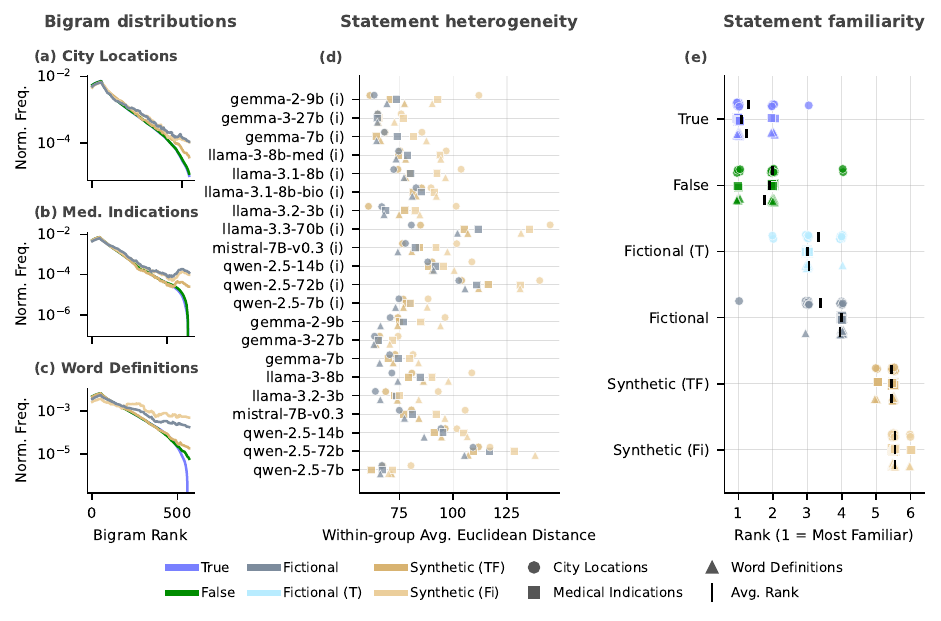}
\caption{
\textbf{Bigram, representational, and token probability analyses of familiarity operationalization.}
Panels \textbf{(a--c)} show rank--frequency curves over character bigrams for City Locations, Medical Indications, and Word Definitions; \textbf{(d)} shows average within-group Euclidean distances between veracity representations for default and instruction-tuned (i) LLMs; and \textbf{(e)} shows familiarity rankings computed from mean object-token probabilities. Lower ranks in \textbf{(e)} indicate greater estimated familiarity. Colors denote statement type: \texttt{True} (purple), \texttt{False} (green), \texttt{Fictional} (dark blue), \texttt{Fictional(T)} (light blue), \texttt{Synthetic (TF)} (dark yellow), and \texttt{Synthetic (Fi)} (light yellow). Marker shapes denote datasets: City Locations (circle), Medical Indications (square), and Word Definitions (triangle). Vertical black bars in \textbf{(e)} denote average ranks across LLMs and datasets. \texttt{Synthetic} statements are consistently less familiar than \texttt{Fictional} statements despite comparable lexical and representational structure, suggesting that the statement types capture familiarity differences beyond low-level artifacts.
}
\label{fig:familiarity}
\end{figure*}

\begin{figure*}[t]
\centering
\includegraphics[width=\textwidth]{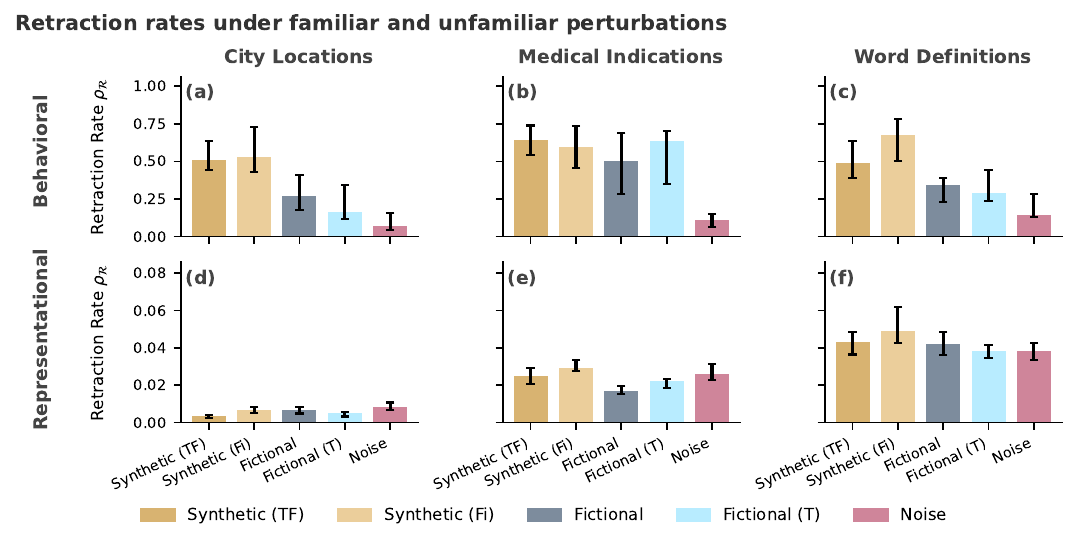}
\caption{
\textbf{Behavioral and representational retraction rates.}
We plot median retraction rates across LLMs for \textbf{(a,d)} City Locations, \textbf{(b,e)} Medical Indications, and \textbf{(c,f)} Word Definitions for \texttt{Synthetic(TF)} (dark yellow), \texttt{Synthetic(Fi)} (light yellow), \texttt{Fictional} (dark blue), \texttt{Fictional(T)} (light blue), and \texttt{Noise} (pink) perturbations. The top row shows behavioral results, while the bottom row shows representational results from \texttt{sAwMIL}. Error bars denote bootstrap confidence intervals. Across datasets, unfamiliar \texttt{Synthetic} perturbations generally induce comparable or greater instability than familiar \texttt{Fictional} perturbations, particularly behaviorally.
}
\label{fig:retractions}
\end{figure*}

\subsection{Exploratory Analyses of Epistemic Retractions}
\label{sec:exp_retraction_structure}

\revision{We investigate whether epistemic retractions exhibit recurring semantic structure across LLMs and datasets. For each $\langle$dataset, LLM$\rangle$ pair, we cluster retracted statement activations using PCA~\cite{pearson1901liii}, UMAP~\cite{mcinnes2018umap}, and HDBSCAN~\cite{campello2013density} with hyperparameters selected via cluster-quality metrics. We then construct contrastive prompts comparing retracted statements against nearby non-retracted statements and use \texttt{Claude Sonnet 4~\cite{claude-sonnet-4}} to generate candidate semantic themes. Additional implementation details appear in Appendix~\ref{appendix:clustering}.}


\section{Results}
\label{sec:results}

\revision{We report (i) bigram, representational, and probabilistic analyses of familiarity, (ii) epistemic stability under perturbations, and (iii) recurring themes in retracted statements.}

\subsection{Evaluating the Familiarity Operationalization}
\label{sec:results:familiarity}

\revision{We first evaluate whether the constructed statement types exhibit the intended distinctions between familiarity and lexical structure (Fig.~\ref{fig:familiarity}). Across datasets, \texttt{Synthetic (TF)} statements approximately replicate the bigram distributions of \texttt{True} and \texttt{False} statements, while \texttt{Synthetic (Fi)} statements approximately preserve the distributions of \texttt{Fictional} statements (Fig.~\ref{fig:familiarity}\textbf{(a--c)}). This reduces the likelihood that downstream stability differences arise solely from lexical statistics.

We next examine representational organization. \texttt{Fictional} statements exhibit similar or lower within-group Euclidean distances than the \texttt{Synthetic} conditions across datasets and LLMs (Fig.~\ref{fig:familiarity}\textbf{(d)}), suggesting \texttt{Fictional} statements are not more heterogeneous than \texttt{Synthetic} ones.

Finally, Figure~\ref{fig:familiarity}\textbf{(e)} estimates familiarity using next-token probabilities as a proxy for pretraining exposure. The average familiarity ordering is \texttt{True}, \texttt{False}, \texttt{Fictional(T)}, \texttt{Fictional}, \texttt{Synthetic(TF)}, and \texttt{Synthetic(Fi)}.\footnote{\revision{LLM-level results are listed in Appendix~\ref{appendix:familiarity}.}} Notably, the \texttt{Synthetic} conditions remain the least familiar despite partially matching the lexical and representational structure of other statement types.}

\begin{table*}[t]
\centering
\resizebox{\textwidth}{!}{%
\begin{tabular}{llll}
\toprule
\textbf{Semantic pattern} & \textbf{Example} & \textbf{Hypothesized source of instability} & \textbf{Observed in} \\
\midrule

Ambiguous referents &
Lima/United States, OH &
Multiple entities/meanings for a term. &
$10/21$ LLMs \\

Rare or obscure concepts &
ptomaine; nard &
Likely infrequent pretraining representation. &
$15/21$ LLMs \\

Borderline semantic relationships &
violin/string; canola/oil &
Indirect or graded semantic relationships. &
$12/21$ LLMs \\

Geopolitical ambiguity &
Dededo Village/Guam &
Competing political categorizations. &
$6/21$ LLMs \\

Technical terminology &
pseudomonas/bacterial genus &
Specialized scientific vocabulary. &
$12/21$ LLMs \\

Qualifier-sensitive biomedical claims &
morphine/colic &
Dependence on subtypes or treatment qualifiers. &
$10/21$ LLMs \\

\bottomrule
\end{tabular}
}
\caption{\textbf{Recurring themes in retracted statements.} We summarize representative semantic themes together with examples, hypothesized sources of instability, and the number of LLMs in which each pattern was observed. Examples are abbreviated object pairs (i.e, ``violin/string'' corresponds to ``A violin is a string.''). Many recurring themes involve ambiguity, weak grounding, specialized terminology, or semantically borderline relationships.}
\label{tab:semantic_retraction_patterns}
\end{table*}

\subsection{Epistemic Stability under Familiar and Unfamiliar Perturbations}
\label{sec:results:stability}

\revision{Behavioral perturbations produce substantial retraction rates across all datasets (Fig.~\ref{fig:retractions}\textbf{(a--c)}). In particular, \texttt{Synthetic} perturbations consistently induce comparable or greater retraction rates than \texttt{Fictional} perturbations. This pattern is strongest for City Locations and Word Definitions (Fig.~\ref{fig:retractions}\textbf{(a,c)}), where median retraction rates under \texttt{Synthetic (TF)}/\texttt{Synthetic (Fi)} are $0.51$/$0.53$ and $0.49$/$0.67$, respectively, compared to $0.27$/$0.17$ and $0.34$/$0.29$ under \texttt{Fictional} and \texttt{Fictional (T)} perturbations. Further, median behavioral retraction rates often exceed $0.5$ ($50\%)$.

Representational perturbations produce much smaller retraction rates (Fig.~\ref{fig:retractions}\textbf{(d--f)}), remaining below $0.02$ for City Locations, below $0.04$ for Medical Indications, and below $0.08$ for Word Definitions. Although the ordering is less consistent across datasets, \texttt{Synthetic} conditions frequently yield comparable or greater instability than \texttt{Fictional} conditions, particularly for Medical Indications and Word Definitions (Fig.~\ref{fig:retractions}\textbf{(e,f)}). Because representational effects are comparatively small, minor absolute differences can change the ordering across perturbation types. Nevertheless, the behavioral and representational results remain qualitatively aligned, suggesting that unfamiliarity induces greater instability in both settings.\footnote{\revision{Appendices~\ref{appendix:llm_level}--\ref{appendix:additional_zs} show that these trends are broadly consistent across LLMs, neighboring layers for \texttt{sAwMIL}, alternative belief contexts, and the \texttt{Mass-Mean} probe. Analogous analyses for epistemic expansions appear in Appendix~\ref{appendix:expansions}.}}}

\subsection{Recurring Themes in Retracted Statements}
\label{sec:results:clustering}

\revision{Retractions frequently concentrate around semantically ambiguous, weakly grounded, or context-dependent claims rather than appearing uniformly across statements (Tab.~\ref{tab:semantic_retraction_patterns}).\footnote{\revision{Domain-level themes are listed in Appendix~\ref{appendix:clustering}.}} Recurring themes include ambiguous referents, rare or technical concepts, qualifier-sensitive biomedical relationships, and semantically graded categorizations.

These themes are broadly consistent with the familiarity-based interpretation developed throughout the paper, as many involve concepts or relationships that are plausibly weakly represented, semantically fragile, or associated with competing associations in pretraining data. Although exploratory, these analyses suggest that instability concentrates in identifiable semantic regions rather than appearing uniformly across statements.}

\section{Discussion}
\label{sec:discussion}

\revision{
This work studies whether the stability of LLM truth judgments depends on the familiarity of the semantic assumptions used to perturb them. Using matched perturbations across representational and behavioral settings, we find that unfamiliar \texttt{Synthetic} statements generally induce greater epistemic instability than familiar \texttt{Fictional} statements, particularly in behavioral evaluations.
}

\revision{
Several findings support this interpretation. First, lexical and representational analyses suggest that stability differences are unlikely to arise solely from lexical and representational differences, as \texttt{Synthetic (TF)} and \texttt{Synthetic (Fi)} preserve the structure of other statement families while remaining comparatively unfamiliar. Second, both representational and behavioral settings exhibit similar trends in retraction rates, with unfamiliar perturbations generally producing equal or greater instability. Finally, clustering analyses suggest that retracted statements are not random but are instead ambiguous, technical, or context-dependent.}

\revision{These findings connect naturally to recent work on robustness and behavioral instability in LLMs. 
Our results suggest that shifts in semantic assumptions can similarly destabilize truth judgments, particularly when the perturbing information is comparatively unfamiliar to the LLM. In this sense, \framework\ 
provides a controlled setting to study how familiarity interacts with representational organization and behavioral robustness.}

\revision{
Beyond robustness evaluation, \framework\ may also help support future work on uncertainty estimation and hallucination analysis. 
Because \framework\ identifies statements most vulnerable to perturbation-induced retractions, it provides a complementary signal to standard accuracy-based evaluations. Stability-based analyses may therefore help identify semantically fragile claims or domains that appear reliable under conventional benchmarks but remain vulnerable under semantic perturbation.
}


\section{Conclusion}
\label{sec:conclusion}

\revision{
We introduce \framework, a framework for evaluating the stability of LLM truth judgments under matched semantic perturbations in representational and behavioral settings. Across $21$ LLMs and three factual domains, unfamiliar \texttt{Synthetic} perturbations generally induce greater instability than familiar \texttt{Fictional} perturbations. Additional lexical, representational, and probabilistic analyses suggest that these effects are not driven solely by low-level artifacts, while clustering analyses reveal recurring themes among retracted statements. Overall, our results suggest that epistemic familiarity is an important factor associated with stability and demonstrate how belief-set analyses can complement traditional factuality and robustness evaluations.
}

\paragraph{AI Usage} 
ChatGPT and Copilot were used to assist with drafting experiment and plotting scripts, code cleaning, and documentation. \revision{Claude was used to generate preliminary labels for the retraction clustering analysis prior to manual review.} All AI-generated content was verified by the authors.


\section{Limitations}

Our perturbations focus on a specific operationalization of epistemic familiarity and a limited set of factual domains. \revision{In particular, familiarity is not directly observable, so we operationalize familiarity through statement construction and likely cultural embeddedness as proxies for pretraining exposure. Although our lexical, representational, and probabilistic analyses help reduce several potential confounds, they cannot fully isolate familiarity from all correlated factors, such as domain coverage or memorization effects.}

Further, while our perturbation design enables controlled comparisons across representational and behavioral settings, it does not exhaustively cover all types of semantic variation. Extending \framework\ to other forms of epistemic ambiguity, such as disputed claims or evolving facts, would further test its generality. \revision{Relatedly, \framework\ evaluates stability under controlled semantic perturbations rather than under naturally occurring conversational dynamics or adversarial attacks observed in deployment settings.} While we emphasize epistemic retractions as a primary signal of instability, other applications may require alternative notions of stability.

Finally, representational analysis considers fixed LLM parameters. Although \framework\ isolates how semantic perturbations interact with existing internal representations, it does not address how belief stability may change in settings where representations themselves evolve over time, such as continual learning or retrieval-augmented systems.


\section{Ethical Considerations}

Our study examines how controlled semantic perturbations can destabilize LLM truth judgments. For example, we find that unfamiliar \texttt{Synthetic} content often induces larger retraction rates than familiar \texttt{Fictional} content. \revision{These findings may help improve robustness evaluation by identifying semantic conditions under which LLM beliefs become unstable.}

\revision{At the same time, perturbation-based analyses could potentially be misused to design prompts or contexts intended to induce instability in deployed systems.} However, our intent is diagnostic rather than exploitative: \framework\ is designed to identify epistemic vulnerabilities in order to support more robust evaluation and model development. We do not propose methods for jailbreaks, persuasion, or targeted belief manipulation, and all experiments are conducted on fixed, open-source LLMs in offline settings.

More broadly, our findings highlight structural limitations of current LLMs rather than providing prescriptions for exploitation. \revision{We therefore view \framework\ primarily as a robustness and reliability evaluation framework that complements traditional accuracy-based benchmarks by analyzing how truth judgments reorganize under controlled semantic variation.} \framework\ is intended solely for research and diagnostic evaluation of pretrained LLMs and is not designed for deployment, belief steering, or real-world decision-making systems.

\FloatBarrier


\section*{Acknowledgments}

We thank Hannes Leitgeb and Branden Fitelson for discussions on $P$-stability and how it might be related to epistemic uncertainty in LLMs. We also thank Zohair Shafi and Moritz Laber for their feedback and discussions on methodological and empirical portions of this work.


\section*{Funding}

This material was sponsored by the Government of the United States under Contract Number FA8702-15-D-0002. The view, opinions, and/or filings contained in this material are those of the author(s) and should not be construed as an official position, policy, or decision of the Government of the United States or Carnegie Mellon University or the Software Engineering Institute unless designated by other documentation.


\section*{Competing interests}

The authors declare no competing interests.


\bibliography{bib.bib}


\clearpage
\FloatBarrier
\begin{appendices}
\onecolumn

\setcounter{figure}{0}
\setcounter{table}{0}
\renewcommand{\thefigure}{A\arabic{figure}}
\renewcommand{\thetable}{A\arabic{table}}


\section{Notation}
\label{appendix:notation}

We summarize the mathematical notation used throughout the manuscript in Table~\ref{appendix:tab:notation}.

\begin{table*}[t]
\centering
\resizebox{0.9\textwidth}{!}{%
\begin{tabular}{ll}
\toprule
\textbf{Symbol} & \textbf{Description} \\
\midrule

$\mathcal{M}$ &
A fixed large language model (LLM). \\

$l$ &
Layer index used for activation extraction. \\

$\mathcal{S} = \{s_i\}_{i=1}^{N}$ &
Set of $N$ natural-language statements. \\

$y_i$ &
Ground-truth veracity label of $s_i$,
$y_i \in \{\texttt{True},\texttt{False},\texttt{Neither}\}$. \\

$\mathcal{O}(s)$ &
Set of object tokens in statement $s$. \\

$t_i$ &
Token at position $i$ in a tokenized statement. \\

$\mathcal{N}$ &
Set of all \texttt{Neither} statements:
$\mathcal{N}=\{s_i\mid y_i=\texttt{Neither}\}$. \\

$\mathcal{N}_{\mathrm{fam}},\ \mathcal{N}_{\mathrm{unf}}$ &
Partition of $\mathcal{N}$ into epistemically familiar (\texttt{Fictional}) and unfamiliar (\texttt{Synthetic}) subsets. \\

$\mathcal{N}_{\mathrm{fam}}^{(T)},\ \mathcal{N}_{\mathrm{fam}}^{(F)}$ &
Canonically true and false subsets of familiar fictional statements. \\

$\phi_{\mathcal M,l}$ &
Representation map from a statement to its layer-$l$ hidden representations under $\mathcal{M}$. \\

$\mathbf{z}_i^{(l)}$ &
Layer-$l$ representation of statement $s_i$,
$\mathbf{z}_i^{(l)}=\phi_{\mathcal M,l}(s_i)$. \\

$\mathcal{D}$ &
Dataset of statements, representations, and labels:
$\mathcal{D}=\{(s_i,\mathbf{z}_i^{(l)},y_i)\}_{i=1}^N$. \\

$\mathcal{D}_{\mathrm{train}},\ \mathcal{D}_{\mathrm{test}}$ &
Training and test splits of $\mathcal{D}$. \\

$\Theta \subseteq \mathcal{S}$ &
Perturbation capturing a semantic assumption specifying which statements are labeled \texttt{True}. \\

$y_i^{\Theta}$ &
Label assigned to statement $s_i$ under perturbation $\Theta$. \\

$\mathcal{N}_\Theta = \Theta \cap \mathcal{N}$ &
Subset of \texttt{Neither} statements relabeled as \texttt{True} under perturbation $\Theta$. \\

$g(\cdot,\Theta)$ &
Evaluation function mapping a statement and perturbation to a binary judgment:
$g(s,\Theta)\in\{\texttt{True},\texttt{Not True}\}$. \\

$h_\Theta$ &
Linear probe trained under labels induced by perturbation $\Theta$. \\

$g_{\mathrm{p}}$ &
Representational instantiation of $g$ via a probe over $\phi_{\mathcal M,l}(s)$. \\

$g_{\mathrm{pr}}$ &
Behavioral instantiation of $g$ via prompting with belief context $C_\Theta$. \\

$C_\Theta$ &
Belief context constructed from statements labeled \texttt{True} under perturbation $\Theta$. \\

$\Theta_0$ &
Baseline perturbation with $\mathcal{N}_{\Theta_0}=\emptyset$
(no \texttt{Neither} relabeled as \texttt{True}). \\

$\mathcal{B}^{\Theta_0}_{\mathrm{true}}$ &
Baseline belief set:
$\{s_i \mid y_i=\texttt{True},\ g(s_i,\Theta_0)=\texttt{True}\}$. \\

$\mathcal{B}^{\Theta_0}_{\mathrm{not \, true}}$ &
Baseline non-belief set:
$\{s_i \mid y_i=\texttt{True},\ g(s_i,\Theta_0)=\texttt{Not True}\}$. \\

$\mathcal{B}^{\Theta}_{\mathrm{true}}$ &
Belief set under perturbed interpretation $\Theta$. \\

$\mathcal{R}$ &
Epistemic retractions:
$\mathcal{R} = \mathcal{B}^{\Theta_0}_{\mathrm{true}} \setminus \mathcal{B}^{\Theta}_{\mathrm{true}}$. \\

$\mathcal{E}$ &
Epistemic expansions:
$\mathcal{E} = \mathcal{B}^{\Theta}_{\mathrm{true}} \setminus \mathcal{B}^{\Theta_0}_{\mathrm{true}}$. \\

$\rho_\mathcal{R}$ &
Retraction rate:
$\rho_\mathcal{R} = |\mathcal{R}| / |\mathcal{B}^{\Theta_0}_{\mathrm{true}}|$. \\

$\rho_\mathcal{E}$ &
Expansion rate:
$\rho_\mathcal{E} = |\mathcal{E}| / |\mathcal{B}^{\Theta_0}_{\mathrm{not \, true}}|$. \\

\bottomrule
\end{tabular}
}
\caption{\textbf{Notation.} Summary of symbols used throughout the manuscript.}
\label{appendix:tab:notation}
\end{table*}


\section{Data}
\label{appendix:data}

\subsection{Data Generation}
\label{appendix:data:data_generation}

We use statements from the City Locations, Medical Indications, and Word Definitions datasets introduced in~\cite{savcisens2025trilemma} under a \texttt{CC-BY-4.0} license. City statements take the form \textit{``The city of [city] is (not) located in [country],''} (omitting \textit{``The city of''} when redundant). Medical statements follow \textit{``[drug] is (not) indicated for the treatment of [disease/condition].''} Word Definition statements draw from three templates: \textit{``[word] is (not) a [instanceOf],''} \textit{``[word] is (not) a type of [typeOf],''} and \textit{``[word] is (not) a synonym of [synonym].''} No personal data, identifying information, or user-generated content is included.

\subsubsection{True and False Statements}
\label{appendix:data:statements}
We take the \texttt{True} and \texttt{False} statements from the datasets introduced in~\cite{savcisens2025trilemma}. All statements are constructed with both affirmative and negated forms.

\subsubsection{Synthetic Statements}
\label{appendix:data:synthetic}

\texttt{Synthetic} entities are generated using a Markov-chain–based name generator (\texttt{namemaker}\footnote{\url{https://github.com/Rickmsd/namemaker}.}) and undergo multi-stage filtering, including database checks, model tagging, and web-search validation, to ensure no accidental overlap with real entities. Validated names are then paired to form grammatically coherent but semantically meaningless statements that follow each template. Because \texttt{Synthetic} entities are unlikely to have appeared in training corpora, LLMs have no basis for assigning them a truth value. Accordingly, these statements function as \texttt{Neither} cases: unknown claims for which belief should be suspended rather than confidently classified as \texttt{True} or \texttt{False}.

\revision{\texttt{Synthetic (TF)} statements, introduced in~\cite{savcisens2025trilemma}, are generated using bigram transition matrices constructed from the \texttt{True} and \texttt{False} statements. Following the same methodology, we additionally construct a new set of \texttt{Synthetic (Fi)} statements using bigram transition matrices derived from the \texttt{Fictional} statements. Together, the \texttt{Synthetic} statements exhibit different surface-form statistics while remaining intentionally unfamiliar to the LLMs, allowing us to partially disentangle epistemic familiarity from lexical structure.}

\subsubsection{Fictional Statements}
\label{appendix:data:fictional}

In addition to \texttt{Synthetic (TF)} statements, which represent \textit{unseen and unknown} claims, we construct new sets of \texttt{Fictional} statements for all three domains. \texttt{Fictional} statements also function as \texttt{Neither} statements in our experiments as they reference entities that do not exist in the real world and therefore lack real-world truth value. However, unlike \texttt{Synthetic (TF)} statements, many \texttt{Fictional} entities are likely to have appeared in LLM training corpora. As such, they represent a complementary form of \texttt{Neither}: claims that an LLM may recognize, but that still lie outside the true–false axis relevant to factual grounding.

To ensure that \texttt{Fictional} statements remain genuinely non-factual, all terms were validated to exclude any real-world overlap, and fictional lexical items appearing in any natural language were excluded to prevent misinterpretation by multilingual LLMs. \texttt{Fictional} statements were then constructed using the same templates as the \texttt{True}, \texttt{False}, and \texttt{Synthetic (TF)} statements, including both affirmative and negated forms.

\paragraph{Fictional City Locations.}
Fictional cities and countries, from ~\cite{wiki_fictional_settlements, wiki_fictional_citystates}, span
literature, film, radio, television, comics, animation, and games. Each $\langle$city, location$\rangle$ pair is included only when an identifiable enclosing region exists. When multiple spatial resolutions are available, we select the most specific (e.g., $\langle$Quahog, Rhode Island$\rangle$ rather than $\langle$Quahog, United States$\rangle$).

\paragraph {Fictional Medical Indications.}
Fictional drug and disease statements are drawn from (1) \textit{NeoEncyclopedia Wiki}~\cite{fandom_fictional_diseases, fandom_fictional_toxins}; (2) ChemEurope’s \textit{List of Fictional Medicines and Drugs}~\cite{chemeurope_fictional_medicines_drugs}; and (3) \textit{The Thackery T. Lambshead Pocket Guide to Eccentric \& Discredited Diseases}~\cite{tomasula2004lambshead}. Drug–disease pairs are included when a treatment relationship exists according to the fictional source.

\paragraph{Fictional Word Definitions.}
Fictional lexical items are compiled from (1) \textit{Gobblefunk}~\cite{beelinguappDahlDictionary}; (2) \textit{Dothraki}~\cite{conlangDothrakiInitial}; and (3) \textit{Na’vi}~\cite{dict_navi_online_dictionary}. Dothraki and Na’vi have formal linguistic structure, whereas Gobblefunk is a playful neologistic extension of English.

\subsubsection{Noise}
\label{appendix:data:noise}

For probing, the \texttt{Noise} statements contain no linguistic content. We generate $n_{\mathrm{noise}} = 0.10 \cdot |\mathcal{D}|$ random activation sequences by sampling from a multivariate Gaussian with per-feature mean, standard deviation, and sequence-length distribution matched to the LLM activations. These distributionally matched but non-semantic sequences allow us to test whether observed representational differences arise from semantic content or from statistical variation in activation space.

For prompting experiments, \texttt{Noise} consists of \texttt{True} statements drawn from domains other than the one under evaluation. This yields belief contexts of identical length and format to other perturbations while removing semantic alignment with the evaluated domain, serving as a behavioral analogue of the non-semantic control used in probing.

\subsection{Data Splits for Probing Experiments}
\label{appendix:data:data_splits}

\begin{table*}[t]
\centering
\resizebox{0.7\textwidth}{!}{%
\begin{tabular}{lcccc}
\toprule
\textbf{Dataset} & \textbf{Train} & \textbf{Calibration} & \textbf{Test} & \textbf{Total} \\
\midrule

City Locations &
$5274$ $(0.54)$ &
$2036$ $(0.21)$ &
$2493$ $(0.25)$ &
$9803$ $(1.00)$ \\

Medical Indications &
$5186$ $(0.54)$ &
$1996$ $(0.21)$ &
$2396$ $(0.25)$ &
$9578$ $(1.00)$ \\

Word Definitions &
$7843$ $(0.53)$ &
$3146$ $(0.21)$ &
$3687$ $(0.25)$ &
$14676$ $(1.00)$ \\

\bottomrule
\end{tabular}
}
\caption{\textbf{Dataset splits.} Number of statements used for training, calibration, and testing. Proportions of the full dataset are reported in parentheses.}
\label{appendix:tab:data_splits}
\end{table*}

Table~\ref{appendix:tab:data_splits} summarizes the partitions used for all experiments. Each dataset is split exclusively into training, calibration, and test sets to prevent data leakage. Approximately $55\%$ of statements are used for training, $20\%$ for calibration, and $25\%$ for testing. We use identical splits in all conditions.


\section{LLMs}
\label{appendix:LLMs}

Table~\ref{appendix:tab:LLMs} lists the twenty-one open-source LLMs used in our experiments. The set spans four major LLM families, Gemma, Llama, Mistral, and Qwen, with approximately $3$ billion \revision{to $72$ billion} parameters and release dates between \revision{February $2024$ and March $2025$.} For each family, we include both base (pre-trained) and chat-tuned variants. Together, these LLMs provide a representative cross-section of current decoder-only architectures varying in scale, origin, and training objectives.

The LLMs are publicly available for research use under their respective licenses (\texttt{Gemma} for Gemma-7b, Gemma-7b-it, Gemma-2-9b, Gemma-2-9b-it, \revision{Gemma-3-27b, and Gemma-3-27b-it}; \texttt{llama3.1} for Llama-3.1-8b and Llama-3.1-8b-Instruct; \texttt{llama3.2} for Llama-3.2-3b, Llama-3.2-3b-Instruct; \texttt{llama3} for Llama3-Med42-8b; \texttt{Bio-Medical-Llama-3-8b LLM License} for Bio-Medical-Llama-3-8b; \revision{\texttt{llama3.3} for Llama-3.3-70B-Instruct;} \texttt{apache-2.0} for Mistral-7B-v0.3, Mistral-7B-Instruct-v0.3, Qwen2.5-7B, Qwen2.5-7B-Instruct, Qwen2.5-14B, and Qwen2.5-14B-Instruct; \revision{\texttt{qwen} for Qwen2.5-72B and Qwen2.5-72B-Instruct}).

\begin{table*}[t]
\centering
\resizebox{\textwidth}{!}{%
\begin{tabular}{llcccccl}
\toprule
\textbf{Official Name} & \textbf{Short Name} & \textbf{Type} & \textbf{\# Decoders} & \textbf{\# Parameters} & \textbf{Primary Layer} & \textbf{Release Date} & \textbf{Source} \\
\midrule

Gemma-$7$b &
\texttt{gemma-7b} &
Base &
$28$ &
$8.54$ B &
C: $14$, M: $19$, W: $17$ &
Feb $21$, $2024$ &
Google \\

Gemma-$2$-$9$b &
\texttt{gemma-2-9b} &
Base &
$42$ &
$9.24$ B &
C: $24$, M: $25$, W: $23$ &
Jun $27$, $2024$ &
Google \\

Gemma-$3$-$27$b-pt &
\texttt{gemma-3-27b} &
Base &
$62$ &
$27.03$ B &
C: $34$, M: $25$, W: $27$ &
Mar $12$, $2025$ &
Google \\

Llama-$3$-$8$b &
\texttt{llama-3.1-8b} &
Base &
$32$ &
$8.03$ B &
C: $18$, M: $17$, W: $17$ &
Jul $23$, $2024$ &
Meta \\

Llama-$3.2$-$3$b &
\texttt{llama-3.2-3b} &
Base &
$28$ &
$3.21$ B &
C: $16$, M: $17$, W: $15$ &
Sep $25$, $2024$ &
Meta \\

Mistral-$7$B-v$0.3$ &
\texttt{mistral-7B-v0.3} &
Base &
$32$ &
$7.25$ B &
C: $18$, M: $17$, W: $18$ &
May $22$, $2024$ &
Mistral AI \\

Qwen$2.5$-$7$B &
\texttt{qwen-2.5-7b} &
Base &
$28$ &
$7.62$ B &
C: $18$, M: $19$, W: $17$ &
Sep $19$, $2024$ &
Alibaba Cloud \\

Qwen$2.5$-$14$B &
\texttt{qwen-2.5-14b} &
Base &
$48$ &
$14.80$ B &
C: $30$, M: $31$, W: $30$ &
Sep $19$, $2024$ &
Alibaba Cloud \\

Qwen$2.5$-$72$B &
\texttt{qwen-2.5-72b} &
Base &
$80$ &
$72.70$ B &
C: $22$, M: $20$, W: $8$ &
Sep $19$, $2024$ &
Alibaba Cloud \\

\midrule

Gemma-$7$b-it &
\texttt{\_gemma-7b} &
Chat &
$28$ &
$8.54$ B &
C: $19$, M: $19$, W: $17$ &
Feb $21$, $2024$ &
Google \\

Gemma-$2$-$9$b-it &
\texttt{\_gemma-2-9b} &
Chat &
$42$ &
$9.24$ B &
C: $27$, M: $26$, W: $25$ &
Jul $27$, $2024$ &
Google \\

Gemma-$3$-$27$b-pt &
\texttt{\_gemma-3-27b} &
Chat &
$62$ &
$27.03$ B &
C: $35$, M: $31$, W: $33$ &
Mar $12$, $2025$ &
Google \\

Llama-$3.2$-$3$b-Instruct &
\texttt{\_llama-3.2-3b} &
Chat &
$28$ &
$3.21$ B &
C: $16$, M: $19$, W: $18$ &
Sep $25$, $2024$ &
Meta \\

Llama-$3.1$-$8$b-Instruct &
\texttt{\_llama-3.1-8b} &
Chat &
$32$ &
$8.03$ B &
C: $18$, M: $19$, W: $18$ &
Jul $23$, $2024$ &
Meta \\

Llama-$3.3$-$70$b-Instruct &
\texttt{\_llama-3.3-70b} &
Chat &
$80$ &
$70.55$ B &
C: $20$, M: $40$, W: $20$ &
Dec $6$, $2024$ &
Meta \\

Llama$3$-Med$42$-$8$b &
\texttt{\_llama-3-8b-med} &
Chat &
$32$ &
$8.03$ B &
C: $18$, M: $16$, W: $15$ &
Aug $12$, $2024$ &
M42 Health \\

Bio-Medical-Llama-$3$-$8$b &
\texttt{\_llama-3-8b-bio} &
Chat &
$32$ &
$8.03$ B &
C: $18$, M: $19$, W: $18$ &
Aug $11$, $2024$ &
Contact Doctor \\

Mistral-$7$b-Instruct-v$0.3$ &
\texttt{\_mistral-7B-v0.3} &
Chat &
$32$ &
$7.25$ B &
C: $19$, M: $21$, W: $18$ &
May $22$, $2024$ &
Mistral AI \\

Qwen$2.5$-$7$B-Instruct &
\texttt{\_qwen-2.5-7b} &
Chat &
$28$ &
$7.62$ B &
C: $19$, M: $21$, W: $18$ &
Sep $19$, $2024$ &
Alibaba Cloud \\

Qwen$2.5$-$14$B-Instruct &
\texttt{\_qwen-2.5-14b} &
Chat &
$48$ &
$14.80$ B &
C: $31$, M: $34$, W: $30$ &
Sep $19$, $2024$ &
Alibaba Cloud \\

Qwen$2.5$-$72$B-Instruct &
\texttt{\_qwen-2.5-72b} &
Chat &
$80$ &
$72.70$ B &
C: $79$, M: $69$, W: $40$ &
Sep $19$, $2024$ &
Alibaba Cloud \\

\bottomrule
\end{tabular}
}
\caption{\textbf{LLMs used in the stability experiments.} We list the official names of the LLMs according to the HuggingFace repository~\cite{wolf2020transformers}. We further specify the shortened name used throughout the paper, whether the model is base or chat-tuned, the number of decoder layers, parameter count, release date, source organization, and the primary veracity layer for the City Locations (C), Medical Indications (M), and Word Definitions (W) datasets according to~\cite{savcisens2025trilemma}. The LLMs are publicly available through HuggingFace~\cite{wolf2020transformers}.}
\label{appendix:tab:LLMs}
\end{table*}


\section{Epistemic Expansions}
\label{appendix:expansions}

\begin{figure*}[t]
\centering
\includegraphics[width=0.88\textwidth]{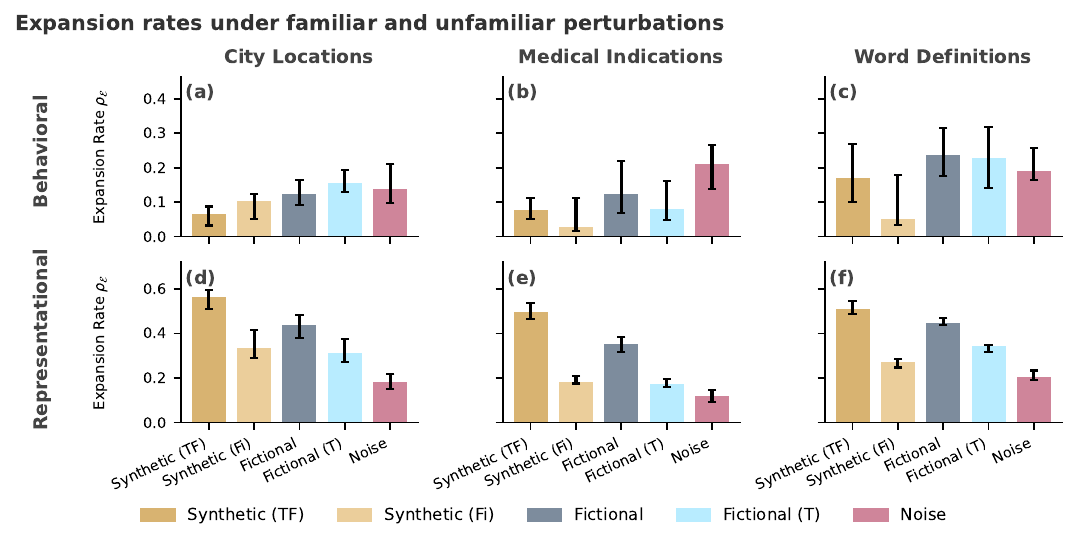}
\caption{ \textbf{Behavioral and representational expansion rates.} We plot median expansion rates across LLMs for \textbf{(a,d)} City Locations, \textbf{(b,e)} Medical Indications, and \textbf{(c,f)} Word Definitions for \texttt{Synthetic(TF)} (dark yellow), \texttt{Synthetic(Fi)} (light yellow), \texttt{Fictional} (dark blue), \texttt{Fictional(T)} (light blue), and \texttt{Noise} (pink) perturbations. The top row shows behavioral results, while the bottom row shows representational results from \texttt{sAwMIL}. Error bars denote bootstrap confidence intervals. \texttt{Synthetic} perturbations tend to induce fewer expansions in the behavioral setting, but \texttt{Synthetic (TF)} consistently produces the highest expansion rates in the representational setting.}
\label{fig:expansions}
\end{figure*}

\revision{The main text primarily focuses on epistemic retractions because they constitute the stronger signal of instability, reflecting previously held beliefs that are withdrawn under perturbation~\cite{leitgeb2014stability}. For completeness, Figure~\ref{fig:expansions} reports expansion rates across LLMs.

In contrast to retractions, behavioral expansions are generally smaller and exhibit weaker separation between perturbation types. In particular, \texttt{Synthetic} perturbations often induce fewer behavioral expansions than \texttt{Fictional} perturbations across datasets. Representational expansions show a partially different trend: \texttt{Synthetic (TF)} frequently produces the highest expansion rates across datasets. These results suggest that epistemic expansions and retractions capture partially distinct aspects of perturbation-induced belief change.}

\begin{figure*}[t]
\centering
\includegraphics[width=0.88\textwidth]{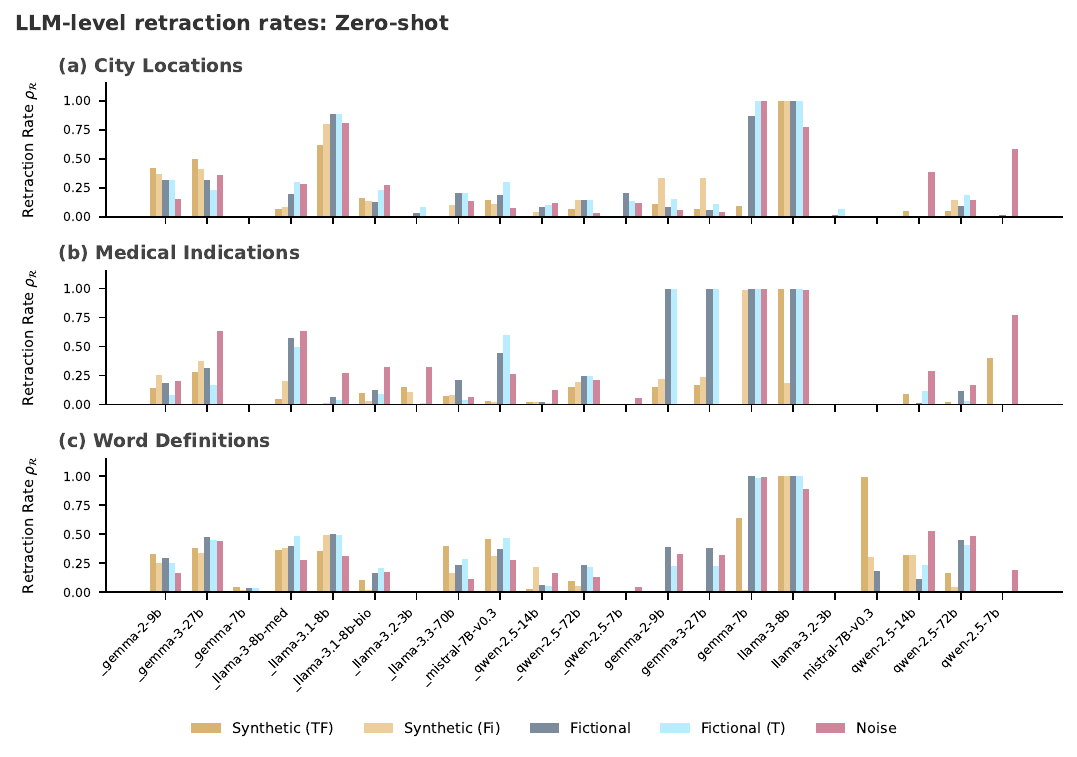}
\caption{\textbf{LLM-level retraction rates for behavioral prompting.}
We show retraction rates on the \textbf{(a)} City Locations, \textbf{(b)} Medical Indications, and \textbf{(c)} Word Definitions datasets for \texttt{Synthetic (TF)} (yellow), \texttt{Synthetic (Fi)} (pale yellow), \texttt{Fictional} (dark blue), \texttt{Fictional (T)} (light blue), and \texttt{Noise} (pink) perturbations for each LLM. The unfamiliar \texttt{Synthetic} perturbations often produce larger behavioral instability than the other perturbations.}
\label{fig:llm_level_zs}
\end{figure*}

\begin{figure*}[t]
\centering
\includegraphics[width=0.88\textwidth]{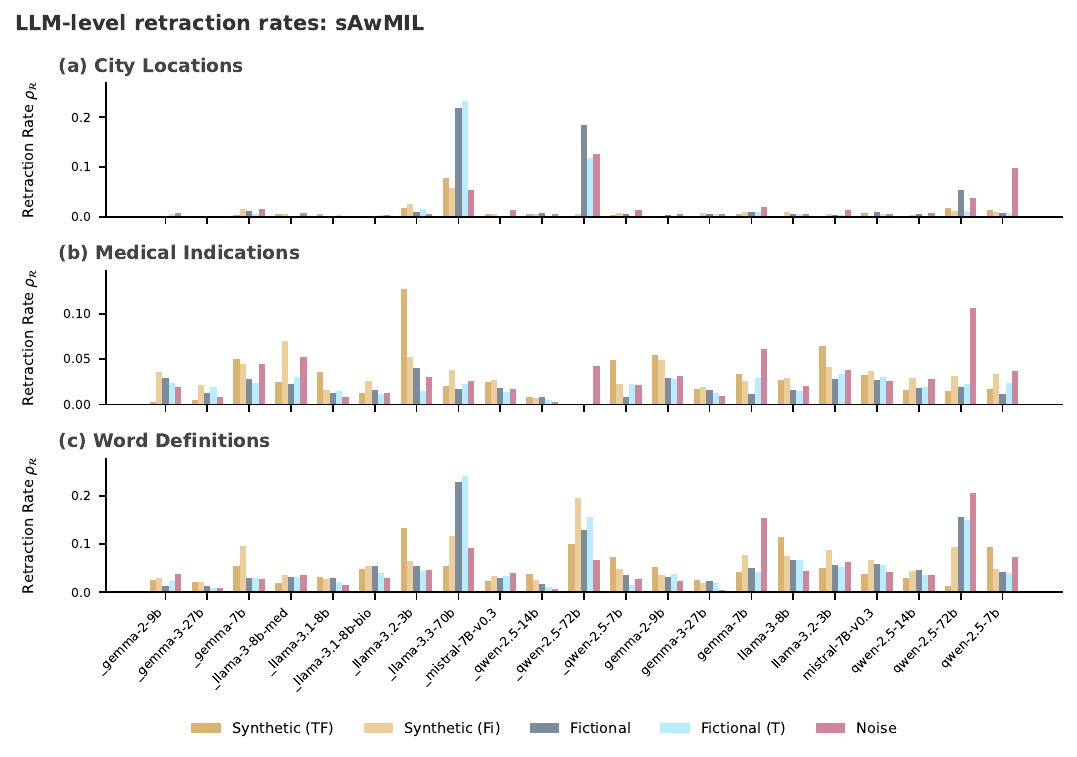}
\caption{\textbf{LLM-level retraction rates for the \texttt{sAwMIL} probe.}
We show retraction rates on the \textbf{(a)} City Locations, \textbf{(b)} Medical Indications, and \textbf{(c)} Word Definitions datasets for \texttt{Synthetic (TF)} (yellow), \texttt{Synthetic (Fi)} (pale yellow), \texttt{Fictional} (dark blue), \texttt{Fictional (T)} (light blue), and \texttt{Noise} (pink) perturbations for each LLM. Retraction rates are almost always under $0.10$, with unfamiliar \texttt{Synthetic} perturbations often leading to more instability than the familiar \texttt{Fictional} ones on Medical Indications and Word Definitions datasets.}
\label{fig:llm_level_sawmil}
\end{figure*}


\section{LLM-level Results}
\label{appendix:llm_level}

\revision{We next present additional LLM-level analyses supporting the aggregate results in Section~\ref{sec:results:stability}. Whereas the main text focuses primarily on median retraction rates across LLMs, the analyses in this appendix illustrate how instability patterns vary across individual LLMs and perturbation conditions.

Figures~\ref{fig:llm_level_zs} and \ref{fig:llm_level_sawmil} visualize retraction rates for each LLM under all perturbation conditions. Across methods, substantial variability exists in absolute instability magnitude, particularly between behavioral and representational settings. Nevertheless, several qualitative patterns repeat consistently across LLM families.

First, unfamiliar \texttt{Synthetic} perturbations frequently induce larger retraction rates than familiar \texttt{Fictional} perturbations. This pattern is especially pronounced in the behavioral setting (Fig.~\ref{fig:llm_level_zs}), where \texttt{Synthetic (TF)} and \texttt{Synthetic (Fi)} often produce among the largest behavioral instability across individual LLMs.

Tables~\ref{tab:aggregate_zero_shot_counts} and \ref{tab:aggregate_sawmil_counts} provide retraction and expansion counts across LLMs. Consistent with the rate-based main-text results, the unfamiliar \texttt{Synthetic} perturbations frequently produce larger retraction counts than familiar \texttt{Fictional} perturbations.}

\begin{table*}[t]
\centering
\includegraphics[width=0.88\textwidth]{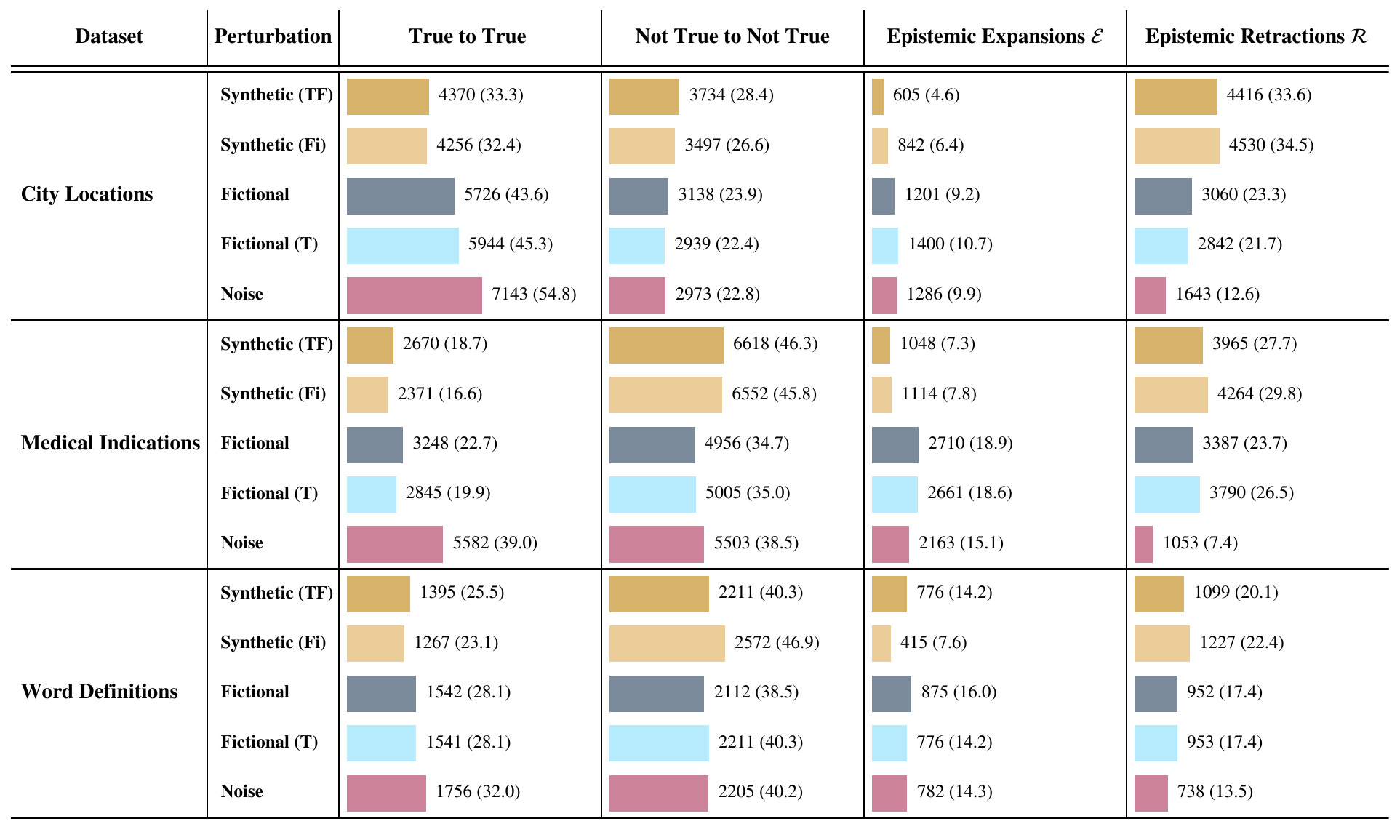}
\caption{\textbf{Epistemic expansions $\mathcal{E}$ and retractions $\mathcal{R}$ for prompting with belief context.}
Counts (percentages) of beliefs that remain stable or undergo expansions $\mathcal{E}$ or retractions $\mathcal{R}$ under \texttt{Synthetic (TF)}, \texttt{Synthetic (Fi)}, \texttt{Fictional}, \texttt{Fictional (T)}, and \texttt{Noise} perturbations, aggregated across all $21$ LLMs. Unfamiliar \texttt{Synthetic} perturbations frequently produce the largest behavioral retraction counts, supporting the conclusion that unfamiliar semantic content acts as a stronger destabilizer of LLM beliefs than familiar \texttt{Fictional} perturbations.}
\label{tab:aggregate_zero_shot_counts}
\end{table*}

\begin{table*}[tb]
\centering
\includegraphics[width=0.88\textwidth]{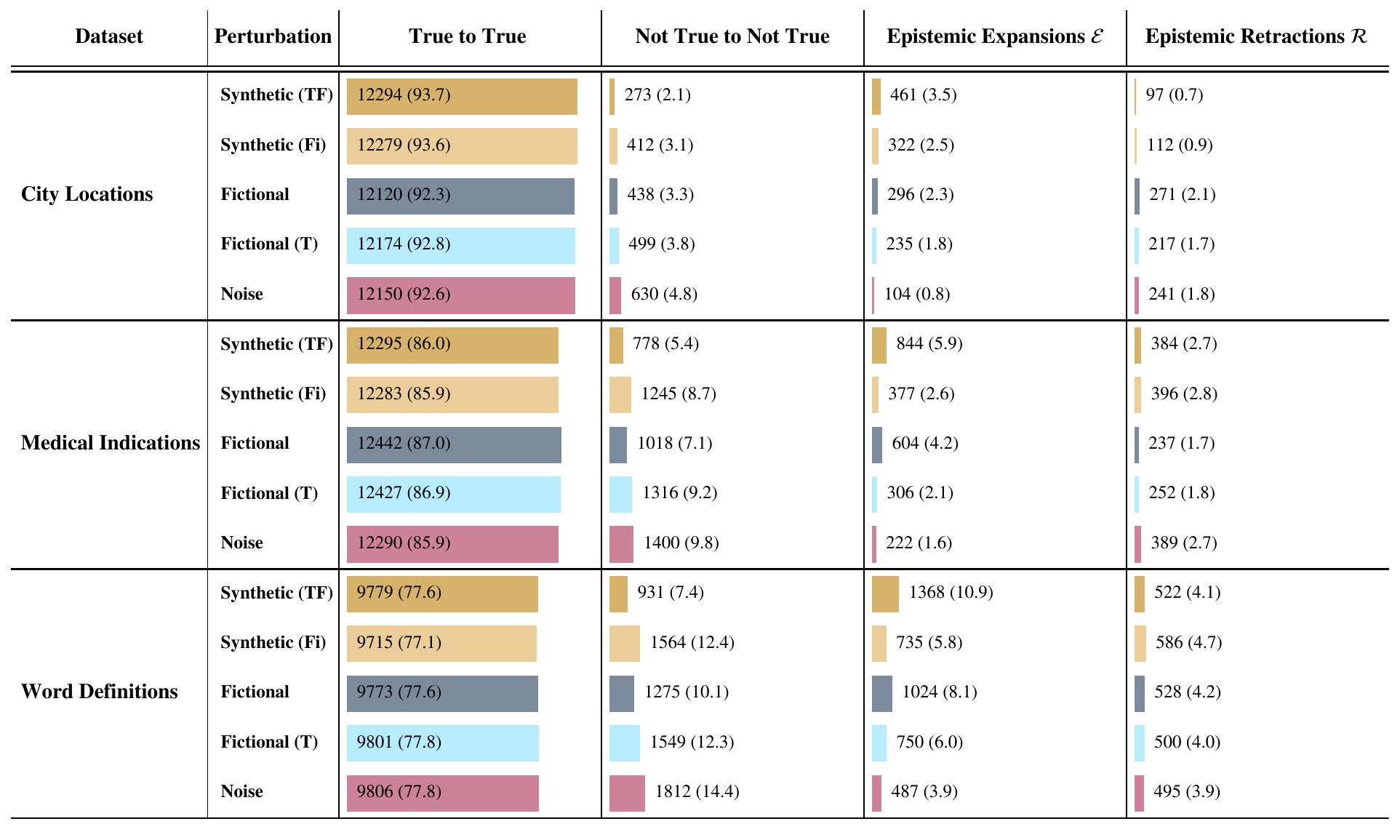}
\caption{\textbf{Epistemic expansions $\mathcal{E}$ and retractions $\mathcal{R}$ for the \texttt{sAwMIL} probe.}
Counts (percentages) of beliefs that remain stable or undergo epistemic expansions $\mathcal{E}$ or epistemic retractions $\mathcal{R}$ under \texttt{Synthetic (TF)}, \texttt{Synthetic (Fi)}, \texttt{Fictional}, \texttt{Fictional (T)}, and \texttt{Noise} perturbations, aggregated across all $21$ LLMs. Although overall retraction rates remain comparatively small, unfamiliar \texttt{Synthetic} perturbations frequently produce comparable or larger instability than familiar \texttt{Fictional} perturbations, particularly for Medical Indications and Word Definitions.}
\label{tab:aggregate_sawmil_counts}
\end{table*}


\section{Layer Robustness Analyses}
\label{appendix:layer_ablation}

\begin{figure*}[t]
\centering
\includegraphics[width=0.9\textwidth]{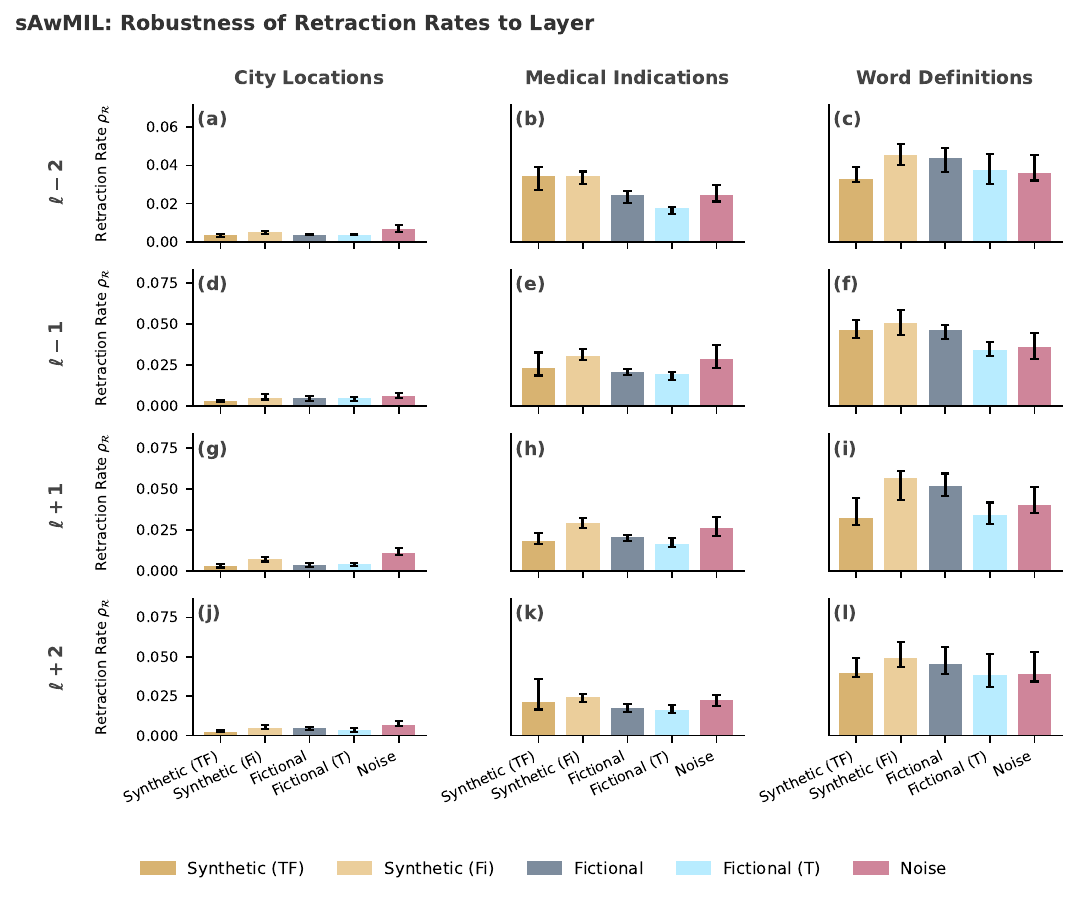}
\caption{
\textbf{sAwMIL layer robustness analyses.}
We plot representational retraction rates across neighboring veracity layers for \textbf{(a--c)} $l-2$, \textbf{(d--f)} $l-1$, \textbf{(g--i)} $l+1$, and \textbf{(j--l)} $l+2$ for City Locations, Medical Indications, and Word Definitions, respectively. Colors indicate perturbation type: \texttt{Synthetic(TF)} (dark yellow), \texttt{Synthetic(Fi)} (light yellow), \texttt{Fictional} (dark blue), \texttt{Fictional(T)} (light blue), and \texttt{Noise} (pink). Error bars denote bootstrap confidence intervals. Retraction patterns remain qualitatively similar across neighboring layers, suggesting that the main-text results are robust to local variation in layer selection.
}
\label{fig:layer_robustness}
\end{figure*}

\revision{To evaluate the robustness of the representational results to the choice of veracity layer, we repeat the \texttt{sAwMIL} analyses on neighboring layers $l-2$, $l-1$, $l+1$, and $l+2$ relative to the layer identified in~\cite{savcisens2025trilemma}. Figure~\ref{fig:layer_robustness} shows that the qualitative ordering of perturbation conditions remains largely unchanged across neighboring layers. In particular, \texttt{Synthetic} perturbations continue to produce comparable or greater retraction rates than \texttt{Fictional} perturbations across most datasets and layers. These results suggest that the main-text findings are not driven by a narrowly tuned choice of representational layer.}


\section{Behavioral Robustness Analyses}
\label{appendix:additional_zs}

\begin{figure*}[t]
\centering
\includegraphics[width=0.9\textwidth]{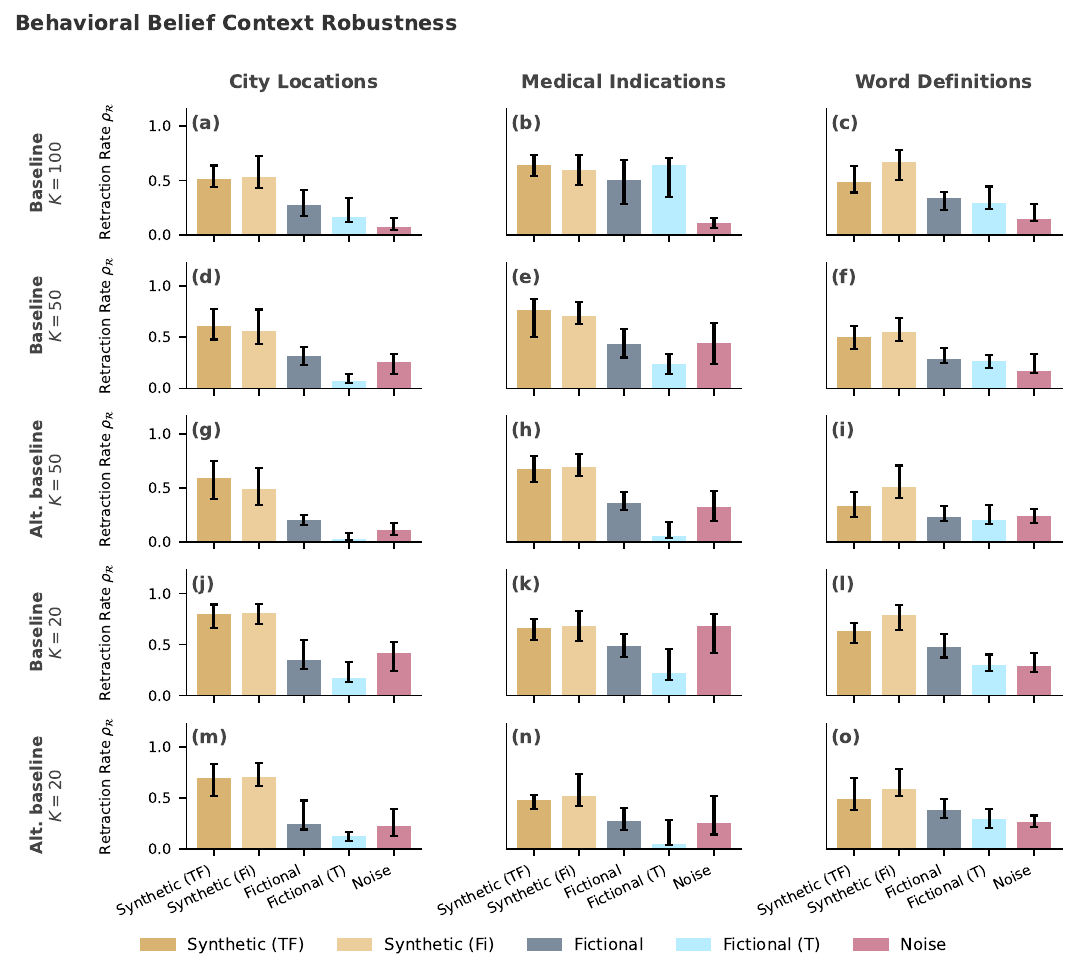}
\caption{
\textbf{Behavioral belief-context robustness analyses.}
We plot behavioral retraction rates under varying belief-context sizes and alternative baselines for \textbf{(a--c)} baseline $K=100$, \textbf{(d--f)} baseline $K=50$, \textbf{(g--i)} alternative baseline $K=50$, \textbf{(j--l)} baseline $K=20$, and \textbf{(m--o)} alternative baseline $K=20$ for City Locations, Medical Indications, and Word Definitions, respectively. Colors indicate perturbation type: \texttt{Synthetic(TF)} (dark yellow), \texttt{Synthetic(Fi)} (light yellow), \texttt{Fictional} (dark blue), \texttt{Fictional(T)} (light blue), and \texttt{Noise} (pink). Error bars denote bootstrap confidence intervals. The qualitative ordering of perturbation conditions remains largely unchanged across belief-context settings and baselines.
}
\label{fig:belief_context_robustness}
\end{figure*}

We additionally evaluate the robustness of the behavioral results to \revision{both belief-context size and} baseline construction. \revision{Specifically, we repeat the behavioral perturbation experiments with smaller belief contexts ($K\in\{20,50\}$) and} with an alternative baseline in which the empty context is replaced by $K$ \texttt{True} training statements. Figure~\ref{fig:belief_context_robustness} shows that the qualitative ordering of perturbation conditions remains largely unchanged across \revision{context sizes} and baselines. These results suggest that the observed instability patterns are driven primarily by the semantic composition of the belief context rather than by its size or mere presence.


\section{Familiarity Analyses}
\label{appendix:familiarity}

\subsection{Additional Methodological Details}

\revision{To empirically characterize familiarity across statement types, we compute token-level next-token probabilities under each LLM using the causal language-model objective. For a tokenized statement
\[
s = (t_1, t_2, \dots, t_n),
\]
the probability assigned to token $t_i$ is
\[
p(t_i \mid t_{<i}),
\]
where $t_{<i}$ denotes all preceding tokens. We compute these probabilities autoregressively for every token in each statement using the LLM's final-layer logits.

Rather than averaging probabilities over all tokens in a statement, we restrict analysis to tokens corresponding to the statement ``objects'' (e.g., \texttt{Paris}, \texttt{asthma}, or \texttt{violin/string}). Concretely, if $\mathcal{O}(s)$ denotes the set of object tokens in statement $s$, we compute the mean object-token probability
\[
\mathrm{Fam}(s)
=
\frac{1}{|\mathcal{O}(s)|}
\sum_{t_i \in \mathcal{O}(s)}
p(t_i \mid t_{<i}).
\]

We focus on object tokens because they contain the primary semantic entities and relationships distinguishing the statement types. Averaging over all tokens would instead be dominated by shared functional words and prompt structure, obscuring familiarity differences associated with the semantically informative portions of the statements.

For each $\langle$LLM, statement type$\rangle$ pair, we then average object-token probabilities across statements to obtain the familiarity estimates reported in Figure~\ref{fig:familiarity}\textbf{(e)} and Tables~\ref{tab:familiarity_cities}--\ref{tab:familiarity_defs}.}

\subsection{LLM-level Familiarity Results}

\begin{table*}[t]
\centering
\small
\resizebox{\textwidth}{!}{%
\begin{tabular}{lcccccc}
\toprule
\textbf{} & \multicolumn{6}{c}{\textbf{Mean object-token probability}} \\
\cmidrule(lr){2-7}
\textbf{LLM} & \texttt{True} & \texttt{False} & \texttt{Fictional} & \texttt{Fictional (T)} & \texttt{Synthetic (TF)} & \texttt{Synthetic (Fi)} \\
\midrule
\_gemma-2-9b & 0.181 & 0.187 & 0.172 & 0.178 & 0.019 & 0.019 \\
\_gemma-3-27b & 0.195 & 0.189 & 0.192 & 0.192 & 0.022 & 0.022 \\
\_gemma-7b & 0.141 & 0.137 & 0.159 & 0.154 & 0.021 & 0.021 \\
\_llama-3-8b-med & 0.237 & 0.224 & 0.175 & 0.180 & 0.020 & 0.020 \\
\_llama-3.1-8b & 0.236 & 0.229 & 0.167 & 0.168 & 0.022 & 0.021 \\
\_llama-3.1-8b-bio & 0.243 & 0.239 & 0.179 & 0.178 & 0.025 & 0.025 \\
\_llama-3.2-3b & 0.216 & 0.252 & 0.169 & 0.165 & 0.020 & 0.020 \\
\_llama-3.3-70b & 0.248 & 0.240 & 0.182 & 0.184 & 0.021 & 0.021 \\
\_mistral-7B-v0.3 & 0.280 & 0.260 & 0.197 & 0.196 & 0.027 & 0.020 \\
\_qwen-2.5-14b & 0.225 & 0.217 & 0.177 & 0.176 & 0.020 & 0.019 \\
\_qwen-2.5-72b & 0.237 & 0.222 & 0.177 & 0.178 & 0.018 & 0.018 \\
\_qwen-2.5-7b & 0.225 & 0.221 & 0.164 & 0.163 & 0.022 & 0.022 \\
gemma-2-9b & 0.172 & 0.167 & 0.154 & 0.154 & 0.022 & 0.022 \\
gemma-3-27b & 0.190 & 0.184 & 0.160 & 0.160 & 0.024 & 0.024 \\
gemma-7b & 0.172 & 0.168 & 0.160 & 0.161 & 0.019 & 0.019 \\
llama-3-8b & 0.229 & 0.220 & 0.160 & 0.160 & 0.023 & 0.023 \\
llama-3.2-3b & 0.219 & 0.219 & 0.148 & 0.146 & 0.021 & 0.021 \\
mistral-7B-v0.3 & 0.282 & 0.267 & 0.202 & 0.202 & 0.026 & 0.019 \\
qwen-2.5-14b & 0.223 & 0.216 & 0.170 & 0.169 & 0.019 & 0.020 \\
qwen-2.5-72b & 0.231 & 0.218 & 0.174 & 0.175 & 0.018 & 0.018 \\
qwen-2.5-7b & 0.217 & 0.218 & 0.162 & 0.160 & 0.021 & 0.021 \\
\bottomrule
\end{tabular}%
}
\caption{\textbf{LLM-level familiarity estimates for City Locations.} We report mean object-token probabilities for each statement type. Higher values indicate greater estimated familiarity. The \texttt{Synthetic} perturbations lead to the lowest mean object-token probability across LLMs.}
\label{tab:familiarity_cities}
\end{table*}

\begin{table*}[t]
\centering
\small
\resizebox{\textwidth}{!}{%
\begin{tabular}{lcccccc}
\toprule
\textbf{} & \multicolumn{6}{c}{\textbf{Mean object-token probability}} \\
\cmidrule(lr){2-7}
\textbf{LLM} & \texttt{True} & \texttt{False} & \texttt{Fictional} & \texttt{Fictional (T)} & \texttt{Synthetic (TF)} & \texttt{Synthetic (Fi)} \\
\midrule
\_gemma-2-9b & 0.320 & 0.309 & 0.055 & 0.109 & 0.019 & 0.019 \\
\_gemma-3-27b & 0.300 & 0.280 & 0.060 & 0.125 & 0.020 & 0.020 \\
\_gemma-7b & 0.279 & 0.274 & 0.053 & 0.115 & 0.021 & 0.021 \\
\_llama-3-8b-med & 0.397 & 0.380 & 0.065 & 0.115 & 0.039 & 0.039 \\
\_llama-3.1-8b & 0.378 & 0.363 & 0.057 & 0.098 & 0.031 & 0.031 \\
\_llama-3.1-8b-bio & 0.400 & 0.389 & 0.065 & 0.116 & 0.041 & 0.041 \\
\_llama-3.2-3b & 0.359 & 0.343 & 0.051 & 0.095 & 0.034 & 0.034 \\
\_llama-3.3-70b & 0.385 & 0.370 & 0.064 & 0.105 & 0.034 & 0.034 \\
\_mistral-7B-v0.3 & 0.432 & 0.416 & 0.088 & 0.176 & 0.058 & 0.031 \\
\_qwen-2.5-14b & 0.377 & 0.367 & 0.066 & 0.112 & 0.035 & 0.035 \\
\_qwen-2.5-72b & 0.399 & 0.403 & 0.081 & 0.126 & 0.040 & 0.041 \\
\_qwen-2.5-7b & 0.371 & 0.365 & 0.055 & 0.104 & 0.037 & 0.037 \\
gemma-2-9b & 0.300 & 0.289 & 0.050 & 0.097 & 0.017 & 0.017 \\
gemma-3-27b & 0.293 & 0.281 & 0.062 & 0.112 & 0.020 & 0.020 \\
gemma-7b & 0.306 & 0.295 & 0.048 & 0.096 & 0.015 & 0.015 \\
llama-3-8b & 0.365 & 0.355 & 0.060 & 0.100 & 0.031 & 0.031 \\
llama-3.2-3b & 0.362 & 0.352 & 0.055 & 0.094 & 0.034 & 0.034 \\
mistral-7B-v0.3 & 0.441 & 0.428 & 0.093 & 0.174 & 0.059 & 0.031 \\
qwen-2.5-14b & 0.375 & 0.366 & 0.062 & 0.107 & 0.036 & 0.036 \\
qwen-2.5-72b & 0.393 & 0.399 & 0.075 & 0.116 & 0.039 & 0.039 \\
qwen-2.5-7b & 0.368 & 0.362 & 0.053 & 0.098 & 0.037 & 0.037 \\
\bottomrule
\end{tabular}%
}
\caption{\textbf{LLM-level familiarity estimates for Medical Indications.} We report mean object-token probabilities for each statement type. Higher values indicate greater estimated familiarity. The \texttt{Synthetic} perturbations lead to the lowest mean object-token probability across LLMs.}
\label{tab:familiarity_med}
\end{table*}

\begin{table*}[t]
\centering
\small
\resizebox{\textwidth}{!}{%
\begin{tabular}{lcccccc}
\toprule
\textbf{} & \multicolumn{6}{c}{\textbf{Mean object-token probability}} \\
\cmidrule(lr){2-7}
\textbf{LLM} & \texttt{True} & \texttt{False} & \texttt{Fictional} & \texttt{Fictional (T)} & \texttt{Synthetic (TF)} & \texttt{Synthetic (Fi)} \\
\midrule
\_gemma-2-9b & 0.083 & 0.079 & 0.034 & 0.035 & 0.003 & 0.003 \\
\_gemma-3-27b & 0.086 & 0.079 & 0.034 & 0.035 & 0.002 & 0.002 \\
\_gemma-7b & 0.091 & 0.090 & 0.041 & 0.044 & 0.003 & 0.003 \\
\_llama-3-8b-med & 0.115 & 0.110 & 0.035 & 0.036 & 0.005 & 0.005 \\
\_llama-3.1-8b & 0.104 & 0.101 & 0.033 & 0.034 & 0.003 & 0.003 \\
\_llama-3.1-8b-bio & 0.116 & 0.115 & 0.037 & 0.038 & 0.006 & 0.005 \\
\_llama-3.2-3b & 0.100 & 0.095 & 0.031 & 0.031 & 0.004 & 0.004 \\
\_llama-3.3-70b & 0.105 & 0.099 & 0.033 & 0.034 & 0.004 & 0.004 \\
\_mistral-7B-v0.3 & 0.144 & 0.139 & 0.054 & 0.057 & 0.007 & 0.006 \\
\_qwen-2.5-14b & 0.110 & 0.111 & 0.035 & 0.036 & 0.004 & 0.004 \\
\_qwen-2.5-72b & 0.128 & 0.124 & 0.036 & 0.036 & 0.005 & 0.005 \\
\_qwen-2.5-7b & 0.110 & 0.110 & 0.035 & 0.035 & 0.004 & 0.004 \\
gemma-2-9b & 0.079 & 0.078 & 0.028 & 0.028 & 0.003 & 0.003 \\
gemma-3-27b & 0.087 & 0.083 & 0.030 & 0.030 & 0.003 & 0.003 \\
gemma-7b & 0.078 & 0.079 & 0.025 & 0.026 & 0.003 & 0.003 \\
llama-3-8b & 0.104 & 0.101 & 0.031 & 0.031 & 0.004 & 0.004 \\
llama-3.2-3b & 0.100 & 0.097 & 0.029 & 0.029 & 0.004 & 0.004 \\
mistral-7B-v0.3 & 0.142 & 0.139 & 0.048 & 0.048 & 0.007 & 0.006 \\
qwen-2.5-14b & 0.109 & 0.109 & 0.032 & 0.033 & 0.004 & 0.004 \\
qwen-2.5-72b & 0.126 & 0.122 & 0.035 & 0.035 & 0.005 & 0.005 \\
qwen-2.5-7b & 0.108 & 0.108 & 0.032 & 0.033 & 0.004 & 0.004 \\
\bottomrule
\end{tabular}%
}
\caption{\textbf{LLM-level familiarity estimates for Word Definitions.} We report mean object-token probabilities for each statement type. Higher values indicate greater estimated familiarity. The \texttt{Synthetic} perturbations lead to the lowest mean object-token probability across LLMs.}
\label{tab:familiarity_defs}
\end{table*}

\revision{Tables~\ref{tab:familiarity_cities}--\ref{tab:familiarity_defs} report the mean object-token probabilities underlying the familiarity rankings shown in Figure~\ref{fig:familiarity}\textbf{(e)}, broken down by individual LLMs and datasets. Higher probabilities correspond to greater estimated familiarity under the autoregressive language-model objective.

Several trends are broadly consistent across datasets and LLMs. First, \texttt{True} and \texttt{False} statements generally receive the highest object-token probabilities. Second, \texttt{Fictional} statements occupy an intermediate regime, with probabilities consistently exceeding those of the \texttt{Synthetic} conditions. Finally, both \texttt{Synthetic (TF)} and \texttt{Synthetic (Fi)} are assigned the lowest probabilities across nearly all evaluated LLMs, despite partially preserving the lexical structure of other statement families.

These results support the interpretation that the constructed \texttt{Synthetic} statements are comparatively unfamiliar to the LLMs while the \texttt{Fictional} statements remain comparatively more familiar due to likely representation in pretraining corpora.}


\section{Retraction Clustering and Semantic Themes}
\label{appendix:clustering}

For each $\langle$dataset, LLM$\rangle$ pair, we identify statements that retract under at least one perturbation condition and extract the corresponding veracity-layer activations. We then perform unsupervised clustering over these activations using a dimensionality-reduction and density-clustering pipeline consisting of principal component analysis (\texttt{PCA},~\cite{pearson1901liii}), Uniform Manifold Approximation and Projection (\texttt{UMAP},~\cite{mcinnes2018umap}), and Hierarchical Density-Based Spatial Clustering of Applications with Noise (\texttt{HDBSCAN},~\cite{campello2013density}). This combination is well-suited for activation-space analyses because it permits irregularly shaped clusters and allows semantically heterogeneous statements to remain unlabeled as noise rather than forcing complete partitioning.

Clustering hyperparameters are selected using unsupervised cluster-quality metrics rather than retraction labels. For each hyperparameter configuration, we compute the silhouette score~\cite{rousseeuw1987silhouettes}, which measures cluster cohesion and separation, together with the fraction of points assigned to noise clusters. We then select the configuration maximizing $\mathrm{Score}=s-0.25\,\rho_{\mathrm{noise}}-\lambda_{\mathrm{tiny}}-\lambda_{\mathrm{few}},$
where $s$ denotes the silhouette score, $\rho_{\mathrm{noise}}$ denotes the fraction of statements assigned to noise, $\lambda_{\mathrm{tiny}}$ penalizes clusters with fewer than three statements, and $\lambda_{\mathrm{few}}$ penalizes solutions with fewer than three clusters. Algorithm~\ref{alg:clustering_pipeline} summarizes the clustering procedure.

\begin{algorithm}[H]
\caption{Retraction Clustering}
\label{alg:clustering_pipeline}
\small
\begin{algorithmic}[1]
\REQUIRE Statements $\mathcal{S}$, perturbation conditions $\Theta$, activations $\phi_{\mathcal M,l}(s)$
\FOR{each $\langle$dataset, LLM$\rangle$ pair}
    \STATE Identify retracted statements and their activations
    \STATE Initialize $\mathrm{Score}^{\star} \leftarrow -\infty$
    \FOR{each clustering hyperparameter configuration $\eta$}
        \STATE Apply PCA dimensionality reduction
        \STATE Apply UMAP embedding
        \STATE Cluster with HDBSCAN
        \STATE Compute silhouette score $s_{\eta}$
        \STATE Compute noise fraction $\rho_{\mathrm{noise},\eta}$
        \IF{a non-noise cluster has $<3$ statements}
            \STATE $\lambda_{\mathrm{tiny},\eta} \leftarrow 0.10$
        \ELSE
            \STATE $\lambda_{\mathrm{tiny},\eta} \leftarrow 0.0$
        \ENDIF
        \IF{there are $<3$ non-noise clusters}
            \STATE $\lambda_{\mathrm{few},\eta} \leftarrow 0.05$
        \ELSE
            \STATE $\lambda_{\mathrm{few},\eta} \leftarrow 0.0$
        \ENDIF
        \STATE $\mathrm{Score}_{\eta}
        \leftarrow
        s_{\eta}
        -
        0.25\,\rho_{\mathrm{noise},\eta}
        -
        \lambda_{\mathrm{tiny},\eta}
        -
        \lambda_{\mathrm{few},\eta}$
        \IF{$\mathrm{Score}_{\eta} > \mathrm{Score}^{\star}$}
            \STATE $\mathrm{Score}^{\star} \leftarrow \mathrm{Score}_{\eta}$
            \STATE $\eta^{\star} \leftarrow \eta$
        \ENDIF
    \ENDFOR
    \STATE Select clustering configuration $\eta^{\star}$
\ENDFOR
\end{algorithmic}
\end{algorithm}

\subsection{Contrastive Prompt Construction}

\revision{To facilitate qualitative interpretation of the resulting clusters, we construct contrastive prompts for \texttt{Claude Sonnet 4} comparing retracted statements against nearby non-retracted statements occupying similar activation-space regions. For each cluster, statements undergoing epistemic retractions are first selected and ranked according to the number of perturbation conditions inducing retractions. We also retrieve nearby non-retracted statements using nearest-neighbor search in standardized activation space with cosine distance.

We then construct structured prompts containing representative retracted and non-retracted statements together with metadata including retraction type, dataset category, and semantic attributes. Prompts explicitly instruct the LLM to focus on recurring noun/entity-level semantic patterns while discouraging reliance on superficial template structure.

For each cluster, \texttt{Claude Sonnet 4} generates (i) a short cluster label, (ii) a description of the shared semantic pattern, (iii) a comparison between retracted and nearby stable statements, and (iv) a hypothesis describing why the cluster may exhibit epistemic instability. Finally, we manually consolidate semantically similar cluster labels into broader recurring themes for analysis in Section~\ref{sec:results:clustering}. We use the following prompt template:
\begin{quote}
\small
You are helping analyze clusters of natural-language statements from LLM activation space.\\[0.5em]

Goal:\\
Identify semantic patterns that may help explain why some statements undergo epistemic retractions while nearby statements do not.\\[0.5em]

Context:\\
- Dataset: [dataset]\\
- Model: [model]\\
- Cluster id: [cluster id]\\
- Cluster size: [cluster size]\\
- Number retracted: [number retracted]\\
- Number nearby non-retracted: [number nearby non-retracted]\\
- Number other baseline statuses: [number other]\\
- Retraction-type counts: [counts]\\[0.5em]

Important instructions:\\
- Do not focus primarily on the surface template.\\
- Do not use the dataset category name as the main explanation unless it captures a more specific pattern.\\
- Focus on shared nouns/entities: ambiguity, polysemy, technicality, concreteness, rarity, familiarity, or semantic domain.\\
- Prefer concise hypotheses grounded directly in the observed nouns/entities.\\
- If there is no clear noun/entity-level pattern, say so rather than forcing one.\\[0.5em]

Retracted examples from this activation-space cluster: [retracted examples]\\[0.5em]

Nearby non-retracted examples: [nearby non-retracted examples]\\[0.5em]

Task:\\
1. Give this cluster a short noun/entity-focused label.\\
2. Describe the pattern shared by the cluster.\\
3. Describe what appears to distinguish the retracted examples from the nearby non-retracted examples.\\
4. Hypothesize why statements involving this kind of content might be vulnerable to epistemic retractions.\\[0.5em]

Return JSON with keys:\\
\quad cluster\_label\\
\quad shared\_cluster\_pattern\\
\quad retracted\_vs\_stable\_contrast\\
\quad retraction\_hypothesis\\
\quad caveats
\end{quote}}

\subsection{Additional Semantic Themes}

\revision{Tables~\ref{tab:cities_loc_semantic_themes}--\ref{tab:defs_semantic_themes} provide additional dataset-specific semantic themes identified through the clustering pipeline described above. Compared to the broader categories presented in the main text (Table~\ref{tab:semantic_retraction_patterns}), these tables preserve finer-grained distinctions that occur within individual datasets and across multiple LLMs.

Consistent with the main-text analysis, many recurring themes involve ambiguity, sparse grounding, unusual semantic relationships, or competing categorizations. However, the dataset-specific breakdown additionally reveals domain-dependent instability patterns. For example, we observe ambiguous place names and geopolitical categorization issues in City Locations, many near-correct or qualifier-sensitive treatment relationships in Medical Indications, and polysemy, near-synonymy, and unusual category relationships in Word Definitions.}

\begin{table*}[t]
\centering
\resizebox{\textwidth}{!}{
\begin{tabular}{llll}
\toprule
\textbf{Theme} & \textbf{Representative examples} & \textbf{Observed in} & \textbf{Related main-text category} \\
\midrule

Ambiguous city names &
Lima/United States, OH &
7 LLMs &
Ambiguous referents \\

Plausible regional pairs &
Rajshahi/Sri Lanka &
6 LLMs &
Borderline semantic relationships \\

Rare or unfamiliar locations &
Elixku/China; Reggane/Algeria &
6 LLMs &
Rare or obscure concepts \\

Geopolitical ambiguity &
Dededo Village/Guam &
6 LLMs &
Geopolitical ambiguity \\

Ambiguous US locations &
Columbia/MD vs Columbia/SC &
3 LLMs &
Ambiguous referents \\

Place vs. non-place confusion &
Jacobo Hunter/Peru &
4 LLMs &
Borderline semantic relationships \\

Uncommon English structure &
Székesfehérvár/Hungary &
4 LLMs &
Rare or obscure concepts \\

\bottomrule
\end{tabular}
}
\caption{\textbf{Additional semantic themes identified for City Locations.}
Representative examples are abbreviated city/location pairs extracted from clustered statements. For example, ``Lima/United States, OH'' corresponds to statements of the form ``The city of Lima is in the United States, OH.'' The ``Observed in'' column reports the number of LLMs in which the corresponding semantic theme appeared, while ``Related main-text category'' links each finer-grained theme to the broader semantic categories summarized in Table~\ref{tab:semantic_retraction_patterns}.}
\label{tab:cities_loc_semantic_themes}
\end{table*}

\begin{table*}[t]
\centering
\resizebox{\textwidth}{!}{
\begin{tabular}{llll}
\toprule
\textbf{Theme} & \textbf{Representative examples} & \textbf{Observed in} & \textbf{Related main-text category} \\
\midrule

Near-correct treatments &
Morphine/colic &
12 LLMs &
Qualifier-sensitive biomedical claims \\

Broad vs. specific conditions &
Doxycycline/bacterial infections &
9 LLMs &
Qualifier-sensitive biomedical claims \\

Experimental drugs &
XL999; ABT-888 &
7 LLMs &
Technical terminology \\

Rare conditions &
Mucormycosis &
8 LLMs &
Rare or obscure concepts \\

Infection-specific treatment &
Vancomycin/lower respiratory tract infections &
5 LLMs &
Qualifier-sensitive biomedical claims \\

Supplement ambiguity &
Aloe vera; Vitamin E &
6 LLMs &
Borderline semantic relationships \\

Treatment vs. side-effect &
Ponatinib/coronary artery disease &
4 LLMs &
Borderline semantic relationships \\

\bottomrule
\end{tabular}
}
\caption{\textbf{Additional semantic themes identified for Medical Indications.}
Representative examples are abbreviated drug--condition pairs extracted from clustered statements. For example, ``Morphine/colic'' corresponds to statements involving morphine and colic-related treatment relationships. The ``Observed in'' column reports the number of LLMs in which the corresponding semantic theme appeared, while ``Related main-text category'' links each finer-grained theme to the broader semantic categories summarized in Table~\ref{tab:semantic_retraction_patterns}.}
\label{tab:med_indications_semantic_themes}
\end{table*}

\begin{table*}[t]
\centering
\resizebox{\textwidth}{!}{
\begin{tabular}{llll}
\toprule
\textbf{Theme} & \textbf{Representative examples} & \textbf{Observed in} & \textbf{Related main-text category} \\
\midrule

Ambiguous proper names &
Johnson/lexicographer; Lewis/author &
10 LLMs &
Ambiguous referents \\

Near-synonyms &
software/package; nuance/shade &
6 LLMs &
Borderline semantic relationships \\

Technical terminology &
pseudomonas/bacteria genus &
18 LLMs &
Technical terminology \\

Broad or unusual categories &
violin/string; canola/oil &
5 LLMs &
Borderline semantic relationships \\

Metaphorical categories &
marksman/shot &
5 LLMs &
Borderline semantic relationships \\

Rare or archaic vocabulary &
nard; rooftree; ptomaine &
19 LLMs &
Rare or obscure concepts \\

Politically sensitive categorizations &
taliban/religious movement &
5 LLMs &
Ambiguous referents \\

Competing valid categorizations &
Rigel/binary vs Rigel/double star &
4 LLMs &
Borderline semantic relationships \\

\bottomrule
\end{tabular}
}
\caption{\textbf{Additional semantic themes identified for Word Definitions.}
Representative examples are abbreviated word--definition or entity--category pairs extracted from clustered statements. For example, ``violin/string'' corresponds to statements involving violin/string semantic relationships. The ``Observed in'' column reports the number of LLMs in which the corresponding semantic theme appeared, while ``Related main-text category'' links each finer-grained theme to the broader semantic categories summarized in Table~\ref{tab:semantic_retraction_patterns}.}
\label{tab:defs_semantic_themes}
\end{table*}


\section{Mass-Mean Results}
\label{appendix:mass-mean}

\begin{figure*}[t]
\centering
\includegraphics[width=0.9\textwidth]{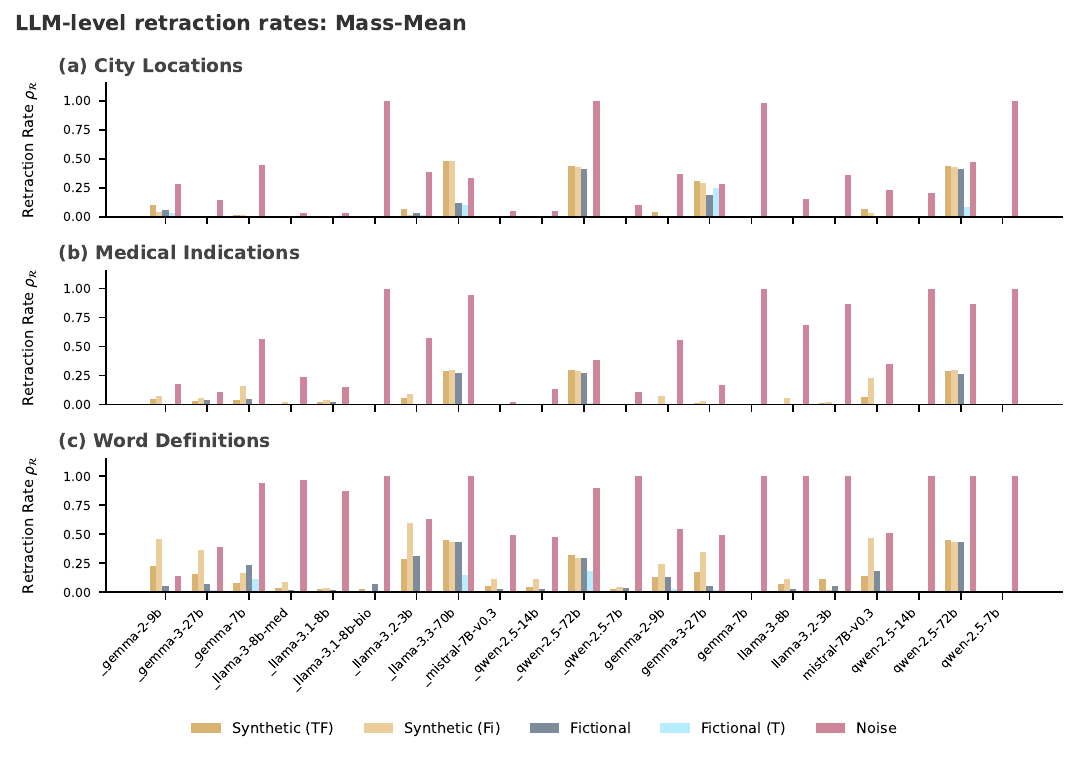}
\caption{\textbf{LLM-level retraction rates for the \texttt{Mass-Mean} probe.}
We show retraction rates on the \textbf{(a)} City Locations, \textbf{(b)} Medical Indications, and \textbf{(c)} Word Definitions datasets for \texttt{Synthetic (TF)} (yellow), \texttt{Synthetic (Fi)} (pale yellow), \texttt{Fictional} (dark blue), \texttt{Fictional (T)} (light blue), and \texttt{Noise} (pink) perturbations for each LLM. While the probe is most susceptible to the \texttt{Noise} perturbation, unfamiliar \texttt{Synthetic} perturbations produce more instability than the familiar \texttt{Fictional} perturbations.}
\label{fig:llm_level_mass_mean}
\end{figure*}

We repeat the representational perturbation experiments using the \texttt{Mass-Mean} probe~\cite{marks2024geometry} to supplement the \texttt{sAwMIL} results. \texttt{Mass-Mean} estimates a ``truth direction'' by taking the vector difference between the centroids of the \texttt{True} and \texttt{False} activations, optionally scaled by the inverse covariance matrix of the data. This approach is inherently sensitive to differences in the centroids and covariance structure of the data, which leads to strong instability when \texttt{Neither} statements are included alongside \texttt{True} and \texttt{False} examples. The \texttt{Mass-Mean} probe shows considerably greater instability compared to \texttt{sAwMIL} (Fig.~\ref{fig:llm_level_mass_mean}), \revision{with retraction rates ranging from about $0.3$ to close to $1.0$ under the noise perturbation. This result is consistent across almost all LLMs (Tab.~\ref{tab:aggregate_mass_mean_counts}).} We note, however, that the \texttt{Synthetic (TF)} \revision{and \texttt{Synthetic (Fi)} perturbations} still produce the most expansions and retractions across domains \revision{if we exclude the \texttt{Noise} perturbation}. We interpret these discrepancies as artifacts of the \texttt{Mass-Mean} probe’s reliance on dataset centroids. This instability, therefore, reflects probe sensitivity rather than genuine representational instability in the LLMs. Accordingly, the \texttt{Mass-Mean} probe is less well suited for quantifying stability within \framework\ than \texttt{sAwMIL}.

\begin{figure*}[t]
\centering
\includegraphics[width=0.9\textwidth]{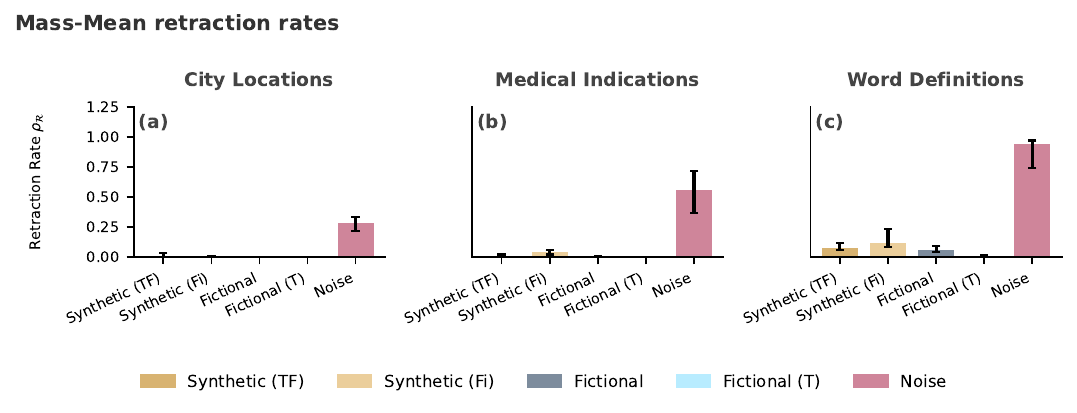}
\caption{
\textbf{\texttt{Mass-Mean} behavioral retraction rates.}
We plot median retraction rates across LLMs for \textbf{(a)} City Locations, \textbf{(b)} Medical Indications, and \textbf{(c)} Word Definitions for \texttt{Synthetic(TF)} (dark yellow), \texttt{Synthetic(Fi)} (light yellow), \texttt{Fictional} (dark blue), \texttt{Fictional(T)} (light blue), and \texttt{Noise} (pink) perturbations using the \texttt{Mass-Mean} probe. Error bars denote bootstrap confidence intervals. Unlike the main-text results, \texttt{Noise} perturbations frequently induce the greatest instability.
}
\label{fig:mass_mean_retractions}
\end{figure*}

\begin{table*}[t]
\centering
\includegraphics[width=0.9\textwidth]{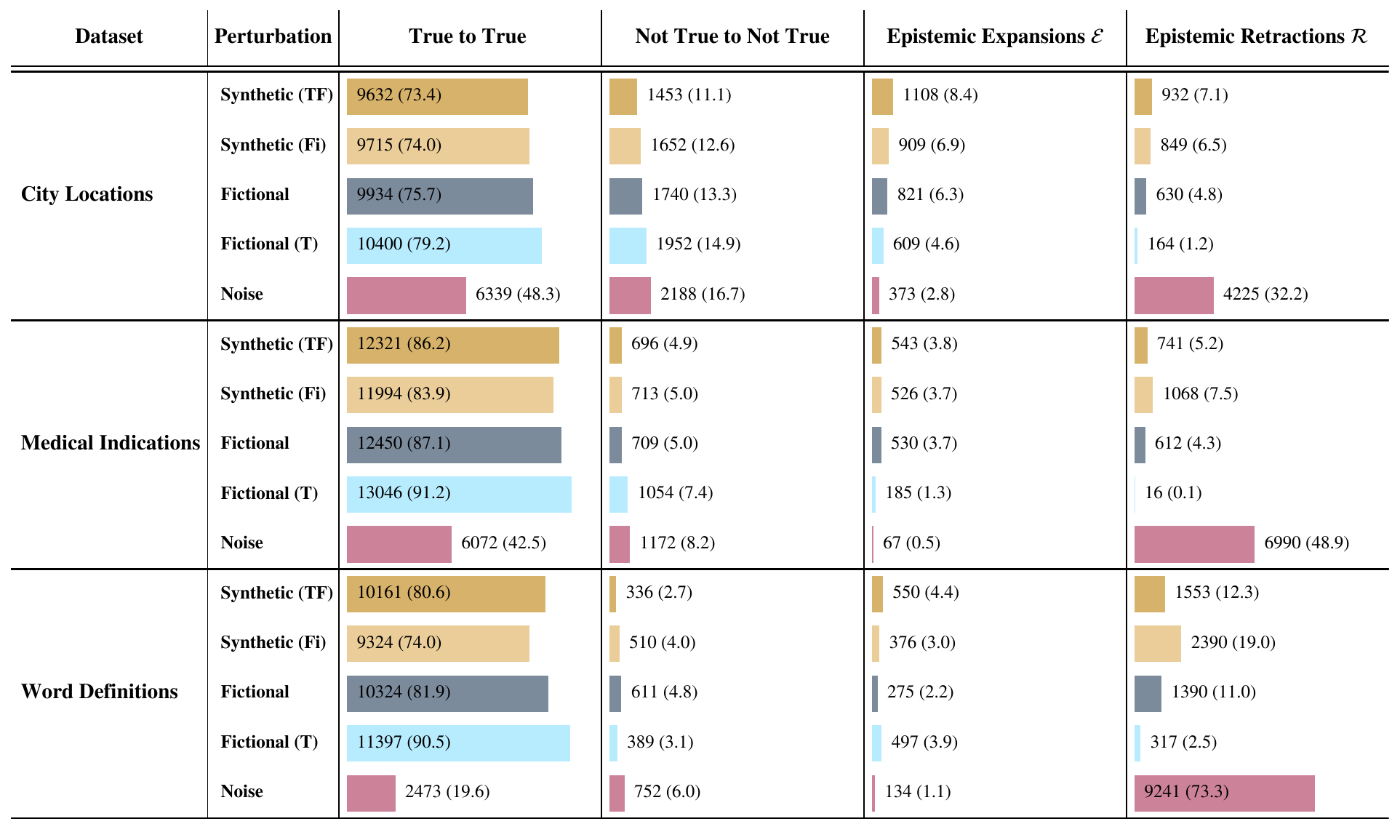}
\caption{\textbf{Epistemic expansions $\mathcal{E}$ and retractions $\mathcal{R}$ for the \texttt{Mass-Mean} probe.}
Counts (percentages) of beliefs that remain stable or undergo epistemic expansions $\mathcal{E}$ or epistemic retractions $\mathcal{R}$ under \texttt{Synthetic (TF)}, \texttt{Synthetic (Fi)}, \texttt{Fictional}, \texttt{Fictional (T)}, and \texttt{Noise} perturbations, aggregated across all $21$ LLMs. Across datasets, unfamiliar synthetic perturbations frequently produce larger retraction counts than familiar fictional perturbations, while \texttt{Noise} induces especially large instability due to the probe's sensitivity to non-semantic perturbations.}
\label{tab:aggregate_mass_mean_counts}
\end{table*}

\end{appendices}

\end{document}